\pdfoutput=1 
\documentclass[peerreview]{IEEEtran}
\usepackage{cite} 
\usepackage{url} 
\usepackage[utf8]{inputenc} 
\usepackage{booktabs} 
\usepackage{graphicx}
\usepackage{color}

\usepackage{algorithm}
\usepackage{algpseudocode}

\begin{document}
\bstctlcite{BSTcontrol}

\title{Applying ranking techniques for estimating influence of Earth variables on temperature forecast error}

 \author{\IEEEauthorblockN{M. Julia Flores\IEEEauthorrefmark{1}, Melissa Ruiz‐Vásquez\IEEEauthorrefmark{2},
 Ana Bastos\IEEEauthorrefmark{2}  and René Orth\IEEEauthorrefmark{2}}

\IEEEauthorblockA{\IEEEauthorrefmark{1} Universidad de Castilla - La Mancha, Spain: Julia.Flores@uclm.es}

\IEEEauthorblockA{\IEEEauthorrefmark{2,3,4} Max-Planck-Institut für Biogeochemie, Germany:
    \\\{mruiz, ana.bastos, rene.orth\}@bgc-jena.mpg.de}
 }

\date{1/3/2024}

\maketitle
\tableofcontents

\IEEEpeerreviewmaketitle
\begin{abstract}
This paper describes how to analyze the influence of Earth system variables on the errors when providing temperature forecasts. The initial framework to get the data has been based on previous research work, which resulted in a very interesting discovery. However, the aforementioned study only worked on individual correlations of the variables with respect to the error. This research work is going to re-use the main ideas but introduce three main novelties: (1) applying a data science approach by a few representative locations; (2) taking advantage of the rankings created by Spearman correlation but enriching them with other metrics looking for a more robust ranking of the variables; (3) evaluation of the methodology by learning random forest models for regression with the distinct experimental variations. The main contribution is the framework that shows how to convert correlations into rankings and combine them into an aggregate ranking. We have carried out experiments on five chosen locations to analyze the behavior of this ranking-based methodology. The results show that the specific performance is dependent on the location and season, which is expected, and that this selection technique works properly with Random Forest models but can also improve simpler regression models such as Bayesian Ridge. This work also contributes with an extensive analysis of the results. We can conclude that this selection based on the top-k ranked variables seems promising for this real problem, and it could also be applied in other domains.

\end{abstract}

\section{Introduction}


Weather forecasting is one of the most valuable services worldwide, given its impact on all scales, from making leisure plans to generating flood containment strategies to preventing problems from extreme temperatures. Numerical weather prediction (NWP) simulates and predicts the atmosphere's behavior over time. NWP techniques involve dividing the atmosphere into a grid system, where mathematical equations represent fundamental physical principles such as atmospheric dynamics, thermodynamics, and fluid motion are applied to each grid point. Those models are highly reliable, and their improvement has been undeniable through the years \cite{WALLACE20061}. There exist multiple and interrelated factors that have contributed to this improvement in weather prediction accuracy over the past few decades: (1) increased computational power, (2) improved observational technologies, and (3) better understanding of atmospheric processes. Besides, the use of ensemble forecasting techniques and the design of better verification and validation practices have led to more reliable predictions and more consistent assessment of forecast performance over time.

The current paper pursues a very specific goal in relation to this topic, and it is a continuation of the work presented in \cite{ruiz2022exploring}. In this study, authors used ecological, hydrological, and meteorological variables to study their potential for explaining temperature forecast errors at the weekly timescale. In particular, Spearman correlations between each considered variable and the forecast error obtained from the European Centre for Medium-Range Weather Forecasts (ECMWF) sub-seasonal-to-seasonal (S2S) forecasts. The data was collected using a thorough procedure that integrated multiple sources, obtaining data across the globe from 2001 to 2017. The findings provided valuable information on processes to improve the temperature forecast skill of the ECMWF–S2S forecasting system.

However, the analysis approach did not consider that Earth system variables are somewhat correlated with
each other, and even if the correlation metric was completely adequate and corrected with the
Benjamini–Hochberg procedure to detect significance, we believe the process could benefit from more sophisticated techniques. The scope of the current paper is more machine-learning oriented, where we will present a framework that should provide an extension that is more robust and informative but also simple enough to be applied to the global context. To provide insight into the variable analysis, we will extract specific locations as use cases among those areas detected as potential at \cite{ruiz2022exploring}. The methodology will cover three main steps: exploratory analysis, constructing an ensemble ranking, and evaluating the results via Ranfom Forest models.



\section{Problem Definition: dataset}

\begin{figure*}[t]
    \centering
    \includegraphics[width=12.5cm]{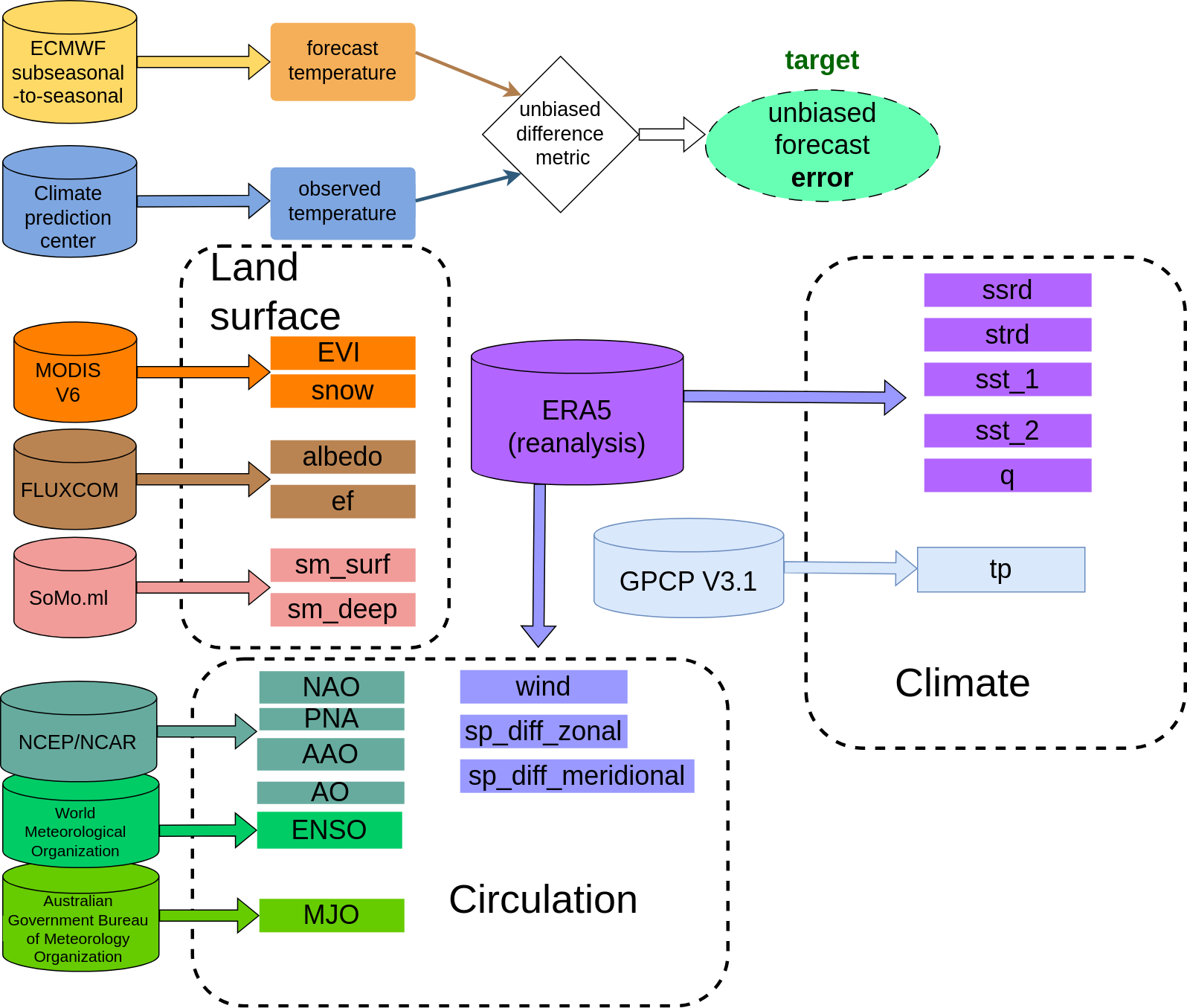}
     \caption{Graphical summary of the data gathered, which involved multiple sources and allows a classification of the feature variables into three main families: Land surface, Climate and Circulation. }
    \label{fig:data-gathering}
\end{figure*}

We have conducted the current research following the same procedure as in \cite{ruiz2022exploring}. The dataset has been constructed integrating distinct sources, which is summarised in the flowchart in Fig. \ref{fig:data-gathering}. These data consider the global land area and the period 2001–2017. Each forecast is run using distinct lead times. Notice that in the current study, we will use the 3-week lead time (15 to 21 days), as it was proved to be the most representative. All the data are registered on a weekly scale. To compute the target variable, error forecast, the Climate Prediction Center (CPC) data were collected and compared with the ECMWF–S2S temperature forecasts. In order to remove bias, this error is the result of de-seasonalizing forecast and reference temperatures at a weekly level. Fig. \ref{fig:data-gathering} shows where the data were obtained from in a schematic way. 

\begin{figure*}[t]
    \centering
    \includegraphics[trim={0 0.5cm 0 0},clip, width=12.5cm]{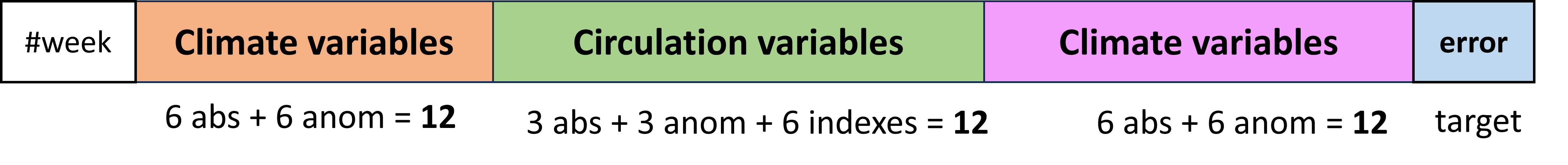}
    \caption{Scheme of a data point referring for a particular week datatime and reflecting information for a cell.}
    \label{fig:data-point}
\end{figure*}

Table \ref{tab:var_description} provides a more detailed description of every variable used in this dataset. It is important to remark that even if there are 21 features, our dataset will include a total of 36, that is because, for those fluctuating variables, we include both the absolute variables (denoted by suffix \textsf{\_abs}) and their anomaly value (denoted by suffix \textsf{\_anom}). This does not apply to the circulation indices ENSO, MJO, NAO, AAO, AO, and PNA, as they are already interpreted as anomalies. When computing the compute anomalies, the long-term trend is subtracted, and the seasonal cycle is also removed on a weekly basis. That, our dataset will contain 36 predictive variables, which accounts for 15 variables  (absolute and anomaly values per each), plus the 6 indices.
already anomalies). An example of data-point is schematized in Fig. \ref{fig:data-point}.

\begin{table}[h!]
    \centering
    \begin{tabular}{|c|p{5.25cm}|}\hline
    Feature name & Brief description \\\hline       
    \textsf{ssrd}&Surface solar radiation downwards \\\hline
    \textsf{strd}&Surface thermal radiation downwards\\\hline
    \textsf{sst\_1}&Sea surface temperature at a grid cell located at the same latitude and 10 degrees eastward of the coastline of the respective grid cell\\\hline
    \textsf{sst\_2}&Sea surface temperature at a grid cell located at the same latitude and 10 degree westward of the coastline of the respective grid cell\\\hline
    \textsf{q}&Specific humidity\\\hline
    \textsf{precip}&Total precipitation\\\hline
    \textsf{wind}&speed wind ERA5 (reanalysis)\\\hline
    \textsf{sp\_diff\_meridional}&Surface pressure differences between 5o north and 5o south of the respective grid cell\\\hline
    \textsf{sp\_diff\_zonal}&Surface pressure differences between 5o east and 5o west of the respective grid cell\\\hline
    \textsf{ENSO}&El Nino Southern Oscillation based on El Nino 3.4 index (anomalies only)\\\hline
    \textsf{MJO}&Madden Julian Oscillation (anomalies only)\\\hline
    \textsf{NAO}&North Atlantic Oscillation (anomalies only)\\\hline
    \textsf{PNA}&Pacific North American pattern (anomalies only)\\\hline
    \textsf{AAO}&Antarctic Oscillation (anomalies only)\\\hline
    \textsf{AO}&Arctic Oscillation (anomalies only)\\\hline
    \textsf{snow}&Snow cover fraction\\\hline
    \textsf{EVI}&Enhanced vegetation index\\\hline
    \textsf{albedo}&Albedo, the fraction of light that a surface reflects\\\hline
    \textsf{EF}&Evaporative fraction, the ratio between latent heat flux and available energy at the land surface\\\hline
    \textsf{SM1}&Surface soil moisture (0 to 10cm)\\\hline
    \textsf{SM\_deep}&Subsurface soil moisture (10 to 50cm)\\\hline
    \end{tabular}\vspace*{0.2cm}
    \caption{Table which shows all the included variables.}
    \label{tab:var_description}
\end{table}

Table \ref{tab:var_description} shows which feature variables have been included in the dataset, with their names as feature (left column) and a brief description. A scheme of the sources from which those variables were measured is shown in Fig. \ref{fig:data-gathering}.

This paper will focus on a Machine Learning approach that intends to extend the methodology followed in \cite{ruiz2022exploring} so that better analysis can be conducted. For this reason, instead of using the global dataset, this work will use a set of specific locations as use cases. We aim to present a methodology that generates more robust rankings but also some validation based on model learning. These locations have been selected with two main criteria: (1) choose those areas where the error forecast had a better margin for improvement, and (2) choose areas in distinct continents. Among the pre-selected locations, we will use just one of them for simplicity in the section on exploratory analysis.

%
%

\section{Correlation Metrics as rankings}

Correlation metrics are a very informative tool in statistical analysis, aiding in understanding relationships between variables. Spearman, Pearson, and Kendall correlation coefficients are among the most widely used metrics for quantifying the degree of association between variables. Each metric offers distinct characteristics, depending on the nature of relationships within datasets.

\subsection{Spearman}
\label{subsec.spearman}

The Spearman correlation coefficient\cite{wissler1905spearman}, denoted by $\rho$ (rho), measures the strength and direction of association between two variables. It is based on the ranks of data points rather than their actual values. It is ideal for assessing monotonic relationships (where the variables move in the same direction but not necessarily linearly). It does not need specific distributional assumptions about the data. Spearman correlation is robust to outliers and is particularly useful when dealing with ordinal or non-normally distributed data. However, it ignores the magnitude of differences between ranks. The formula for Spearman’s $\rho$ involves calculating the difference between the ranks of paired data points (see Eq. \ref{eq:spearman}).

\begin{equation}
    \rho = 1 - \frac{6 \sum d_i^2}{n(n^2 - 1)}
\label{eq:spearman}    
\end{equation}
Where:
\begin{itemize}
    \item \(\rho\) is the Spearman correlation coefficient.
    \item \(d_i\) is the difference between the ranks of corresponding paired observations.
    \item \(n\) is the number of paired observations.
\end{itemize}

\subsection{Pearson}

The Pearson correlation coefficient \cite{bollen1981pearson}, often represented by $r$, quantifies the linear relationship between two variables. It measures the strength and direction of the linear association between variables, assuming that the data follow a bivariate normal distribution. It is especially suitable for assessing linear associations. Pearson correlation is sensitive to outliers and deviations from linearity but provides information about both the strength and direction of the linear relationships within the data. The formula for Pearson’s $r$ involves covariance and standard deviations of the variables. (see Eq. \ref{eq:pearson}).

\begin{equation}
    r = \frac{\sum (x_i - \bar{x})(y_i - \bar{y})}{\sqrt{\sum (x_i - \bar{x})^2 \sum (y_i - \bar{y})^2}}
\label{eq:pearson}\end{equation}
Where:
\begin{itemize}
    \item \(r\) is the Pearson correlation coefficient.
    \item \(x_i\) and \(y_i\) are the individual data points.
    \item \(\bar{x}\) and \(\bar{y}\) are the means of the x and y variables, respectively.
\end{itemize}

\subsection{Kendall}

The Kendall Tau\cite{kendall} correlation coefficient, denoted by $\tau$ (tau), assesses the strength of the ordinal association between two variables. It compares the concordance and discordance of the ranks between paired observations, irrespective of the actual values of the variables. It is particularly useful when dealing with ordinal data or ranks. Kendall's correlation is robust to outliers and non-normally distributed data, making it suitable for analyzing relationships in ranked data sets or when the assumptions of linearity are not met. However, it ignores the magnitude of differences between ranks. The formula for Kendall’s correlation involves counting concordant and discordant pairs (see Eq. \ref{eq:kendall}).

\begin{equation}
\tau = \frac{{\#{Concordant\_pairs}} - {\#{Discordant\_pairs}}}{{\frac{1}{2}n(n-1)}}
\label{eq:kendall}\end{equation}
Where:
\begin{itemize}
    \item \(\tau\) is the Kendall Tau correlation coefficient.
    \item \(n\) is the number of observations.
    \item $\#{Concordant\_pairs}$: number of concordant pairs.
    \item $\#{Discordant\_pairs}$: number of discordant pairs.
\end{itemize}

A pair of data points is considered \textbf{concordant} if their order remains consistent when comparing the two variables. In other words, if both variables show the same relative ranking for that pair.
For example, if $x_i$ and $y_i$ are data points, they form a concordant pair if $(x_i > x_j)$ and $(y_i > y_j)$ or $(x_i < x_j)$ and $(y_i < y_j)$. On the contrary, a pair of data points is considered discordant if their order differs between the two variables, that is, if one variable ranks them differently from the other. For example, if $x_i$ and $y_i$ are data points, they form a discordant pair if $(x_i > x_j)$ and $(y_i < y_j)$ or $(x_i < x_j)$ and $(y_i > y_j)$. 

The value of Kendall’s Tau ranges from -1 to 1:
\begin{itemize}
\item $\tau = -1$: Perfect negative association (all pairs are discordant).
\item $\tau = 0$: No association (random arrangement of pairs).
\item $\tau = 1$: Perfect positive association (all pairs are concordant).
\end{itemize}

\subsection{Correlation as a ranking}

The general recommendation for choosing one of them is to observe the data type, the research question to be answered, and the assumptions. Each metric provides valuable insights into the nature of associations, allowing researchers to make informed decisions in scientific investigations. That is why the base paper \cite{ruiz2022exploring} used Spearman correlation, as the study looked for potentially nonlinear relationships. In this paper, we will show how these correlations are also valuable in a way that the exact values are not strictly utilized, but what matters is their order or rank. That is, we are going to use them as rankings. As there is extensive research on ranking combinations, we will explore how to make a consensus ranking using the three of them and pursue a more robust ordering of the feature variables with respect to the target (error forecast).

When generating a ranking of features with respect to a target variable, each correlation metric offers unique advantages. Spearman correlation is suitable for identifying monotonic relationships, making it valuable for ordinal data or when the relationship between variables is non-linear. Pearson correlation, on the other hand, is ideal for detecting linear associations but may be sensitive to outliers. Kendall Tau correlation provides insights into ordinal associations, making it useful for ranked data sets. Our first contribution is a framework where we will generate a combined ranking that will work as a more complete and general method using an ensemble-based approach. In \cite{caruana2004ensemble} the authors explored the benefits of ensemble methods in a broader context, highlighting their ability to improve predictive accuracy and robustness. Ensemble methods leverage the wisdom of crowds by combining multiple base models, each capturing different aspects of the data, resulting in more reliable predictions. This collective intelligence enables ensemble methods to achieve superior performance over individual models. This applies to models but could also be used for any information we are inferring from data, such as rankings of variables with respect to the target variable, which is our case.

\section{Machine Learning tools}

\subsection{Ranking techniques}

Ranking methods are pivotal in various domains where disparate factors or labels necessitate ordered arrangements. Different raters or judges may assign distinct rankings to the same set of items in numerous scenarios, such as evaluating preferences, assessing performances, or aggregating opinions. This theoretical problem not only underscores the subjectivity inherent in human judgments but also poses computational challenges in reconciling diverse rankings into a cohesive and representative order. From a computational perspective, the challenge of finding a consensus between distinct rankings involves establishing a framework that quantifies the disparity between individual rankings and facilitates the synthesis of a unified ranking. 

Let $N$ denote the number of items being ranked, and let $R_i$ represent the ranking assigned by a particular rater or judge $i$, where $i \in \{1, 2, \dots, M\}$, where $M$ is the total number of raters or judges. Each ranking $R_i$
  comprises a permutation of the items, reflecting the preferences or assessments of the respective rater. Formally, we can represent a ranking $R_i$ as a permutation vector $\pi_i = (\pi_{i1}, \pi_{i2}, ..., \pi_{iN})$, where 
$\pi_{ij}$  denotes the position of the $j$-th item in the ranking assigned by rater $i$. The challenge lies in devising methodologies to compare and reconcile these permutation vectors to generate a consensus ranking that accurately reflects the collective preferences or assessments.

One widely employed method for combining rankings is the Borda count \cite{Saari2023}, which assigns points to each item based on its position and sums them across all rankings to determine the aggregate ranking. The Borda score ($B_j$) is calculated as: $B_j = \sum_{i=1}^{M} (N - \pi_{ij} + 1)$. This method is simple but works properly in the general case, and it can be easily applied to generate an ensemble ranking.

\subsection{Decision Trees and ensemble models}

A decision tree is a fundamental building block in machine learning. It can be seen as a flowchart where each internal node represents a decision based on a feature, and each leaf node corresponds to an outcome: a class label for classification or a numerical value for regression. Decision trees are intuitive, interpretable, and capable of handling both categorical and continuous features. Decision trees have leaf nodes, which represent the prediction, and intermediate nodes, which could be as a decision about the path to be taken. Therefore, when constructing a decision tree, there are two main mechanisms:
\begin{itemize}
\item \textbf{Split}: The tree recursively splits the dataset into subsets based on the most informative feature. The goal is to minimize impurity (e.g., variance, Gini index, or entropy) within each subset.
\item \textbf{Determine the prediction}: At each leaf node, the model predicts the target value by averaging the training samples within that leaf (for regression) or by majority voting (for classification).
\end{itemize}

\begin{figure}
    \begin{tabular}{cc}
    \includegraphics[width=4.5cm]{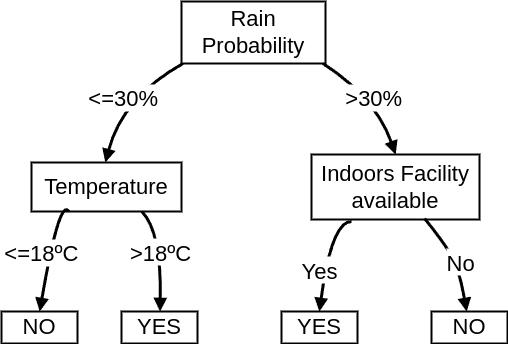} & \includegraphics[width=2.5cm]{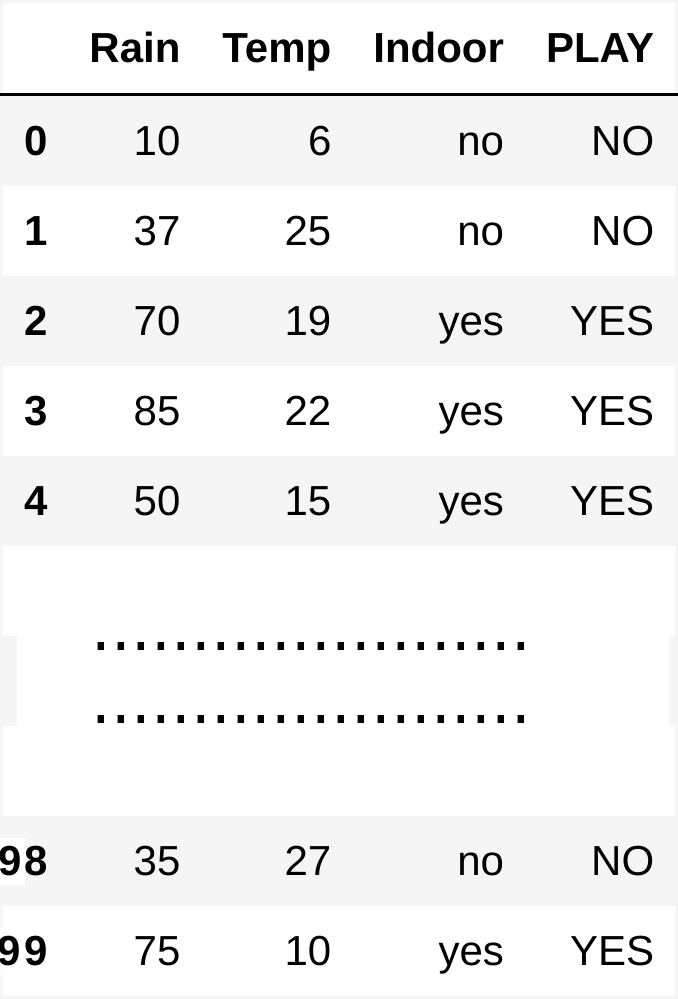}
    \end{tabular}
    \caption{Example Decision Tree (left). Partial and schematic dataset from which it could have been inferred (right).}
    \label{fig:dt-example}
\end{figure}

As an illustration, Fig. \ref{fig:dt-example} shows a hand-crafted and simple Decision Tree based on the classic \emph{play tennis} example \cite{mitchell1997machine}. In this case it is for classification, as the target variable has categorical labels. This shows three intermediate decision nodes that correspond with three feature variables. Rain probability and temperature are continuous variables, and in this case a binary threshold determines the decision. Indoors Facility is a binary categorical variable (Yes/No). On the right, some cases of the possible input dataset is shown. There are multiple algorithms for learning a decision tree, as ID3 \cite{quinlan1986induction} or C4.5 \cite{quinlan2014c4}. Once we have it constructed, their use for inference or prediction is simple, as any possible case can only follow one branch that will direct for one final decision in the corresponding leaf node. For instance, in a new case which could be \{Rain probability=7\%, Temperature=$27^oC$, Indoors=no\}, the prediction would be 'YES' because the path leads to that leaf node.

Decision trees are versatile machine-learning models for classification and regression tasks. They mimic human decision-making processes by dividing the input space into regions and making decisions based on the features present in those regions. Each internal node of the tree represents a decision based on a feature, and each leaf node represents the predicted outcome. Decision trees are easy to interpret and understand, making them popular choices for various applications. However, regular decision trees tend to overfit due to their high flexibility in partitioning the feature space to fit the training data precisely. As decision trees grow deeper, they become increasingly complex and can capture noise or outliers present in the training data, leading to poor generalization performance on unseen data. Overfitting occurs when the model learns to memorize the training data rather than capturing underlying patterns or relationships.

\subsection{Ensemble models: Random Forest}
\label{subsec:rf}

Ensemble methods have been extensively studied and empirically validated to outperform individual decision trees. In \cite{dietterich2000ensemble} the effectiveness of ensemble methods, particularly boosting and bagging, was demonstrated across different datasets. This study concluded that ensemble methods consistently yielded lower error rates and better generalization performance than individual classifiers, including decision trees. In particular, Random Forest is recognised as one of the most widely used ensemble methods. There are several key factors contributing to its success. One significant factor is its robustness against overfitting, which is achieved through the combination of multiple decision trees trained on random subsets of the data and features. By introducing randomness during both sample selection and feature selection, Random Forest effectively decorrelates the individual trees, reducing the risk of overfitting and improving generalization performance \cite{breiman2001random}. Moreover, Random Forest exhibits remarkable flexibility and scalability, making it suitable for various applications across various domains. The algorithm can handle high-dimensional data with categorical and numerical features without requiring extensive preprocessing, simplifying the modeling process and reducing the risk of information loss. This versatility has contributed to the widespread adoption of Random Forest in real-world scenarios where complex data structures and large datasets are common \cite{liaw2002classification}. The interpretability of Random Forest models is another advantageous feature that has contributed to its popularity, and this is a perspective that we will discuss in the explainability section. Finally, Random Forest's computational efficiency and ease of implementation have made it accessible to both researchers and practitioners. All the listed characteristics have encouraged us to use this model, and its prediction, for analysis and validation purposes throughout the experimentation.

\begin{algorithm}
\caption{Random Forest Algorithm}
\begin{algorithmic}[1]
\State Initialize an empty forest (a collection of decision trees).
\For{each tree}
    \State Randomly sample the training data with replacement (bootstrap).
    \State Train a decision tree on the sampled data and randomly select a subset of features for each split.
    \State Add the trained tree to the forest.
\EndFor
\end{algorithmic}\label{alg:RF}
\end{algorithm}

Once a Random Forest model has been learned (see steps at Alg. \ref{alg:RF}), it can be easily used for prediction by averaging the predictions from all trees (for regression), or using the majority vote (for classification).

One utility that makes Random Forest particularly interesting for our study is its ability to generate feature importance. They allow the computation of the relative importance of each feature (or variable) in predicting the target variable. It is particularly useful for understanding which features contribute the most to the predictive power of the model.

In exploratory analysis, this capability is very commonly used for identifying the predictive power by examining feature importance scores, as one can identify which features have the most influence on the model's predictions. This helps you understand which aspects of your data are most relevant in explaining the target variable. Also, this is applied to feature subset selection by focusing on the most informative features and potentially discarding less important ones. This can simplify models and improve computational efficiency. Also, as high feature importance may suggest strong relationships between the feature and the target variable, exploring these relationships further can provide insights into the underlying dynamics of the data.

\subsection{Explainability}
\label{subsec:explainability}

While ensemble methods are often perceived as black-box models, Random Forest offers insights into feature importance, allowing users to understand the relative contribution of each feature to the model's predictions. This interpretability is crucial in domains where model transparency and explainability are paramount, such as healthcare and finance \cite{louppe2014understanding}. Feature importance can help users understand the factors driving the model's decisions. Feature importance in Random Forests is typically calculated based on how much each feature decreases impurity across all decision trees in the forest. Features that consistently contribute to reducing impurity are deemed more important in determining the final predictions. This information allows stakeholders to identify which features have the most significant influence on the model's output, facilitating trust and transparency in decision-making processes. We will use this utility in our analysis. Also, as we aimed to determine those variables that most influenced forecast error, we can also apply specific explainability techniques that account for those contributions. Two popular approaches for interpreting Random Forest models and other complex machine learning models are SHAP  and LIME.

\subsubsection{SHapley Additive exPlanations}

SHAP \cite{lundberg2017unified} values provide a theoretically grounded framework for explaining individual predictions by quantifying the contribution of each feature to the model's output. By leveraging concepts from cooperative game theory, SHAP values assign a unique contribution to each feature, taking into account interactions between features. This approach offers a holistic understanding of how each feature influences a specific prediction, enhancing the interpretability of Random Forest models.

SHAP values for Random Forest involve decomposing the model’s prediction into contributions from individual features. For each instance, we compute the difference between the expected prediction (average over all instances) and the actual prediction when a specific feature is included. Their interpretation is:
\begin{itemize}
\item Positive SHAP value: The feature contributes positively to the prediction.
\item Negative SHAP value: The feature has a negative impact on the prediction.
\item Sum of SHAP values equals the difference between the actual prediction and the expected prediction.   
\end{itemize}

The SHAP Summary Plot shows the SHAP values for each feature across the entire dataset. Notice that features with larger absolute SHAP values contribute more significantly to predictions. We could also observe particular predictions, and then look at the individual SHAP Values for a specific instance. This allows the SHAP values to be examined to understand which features influenced the prediction.

\subsubsection{Local Interpretable Model-agnostic Explanations}

LIME \cite{ribeiro2016should} focuses on providing local explanations for individual predictions by approximating the behavior of complex models with simple, interpretable models. It generates locally faithful explanations by perturbing input instances and observing changes in predictions, allowing users to understand how small changes in input features affect the model's output. LIME's model-agnostic nature makes it applicable to a wide range of machine learning models, including Random Forest, without requiring knowledge of the underlying model architecture. LIME provides feature importance coefficients for the local model, then the values allows the interpretation of the coefficients to understand the impact of features on one particular prediction.

\section{Methodology}

\subsection{Chosen Locations}
\label{subsec:chosen_locations}

\begin{figure}[t]
    \includegraphics[width=9cm]{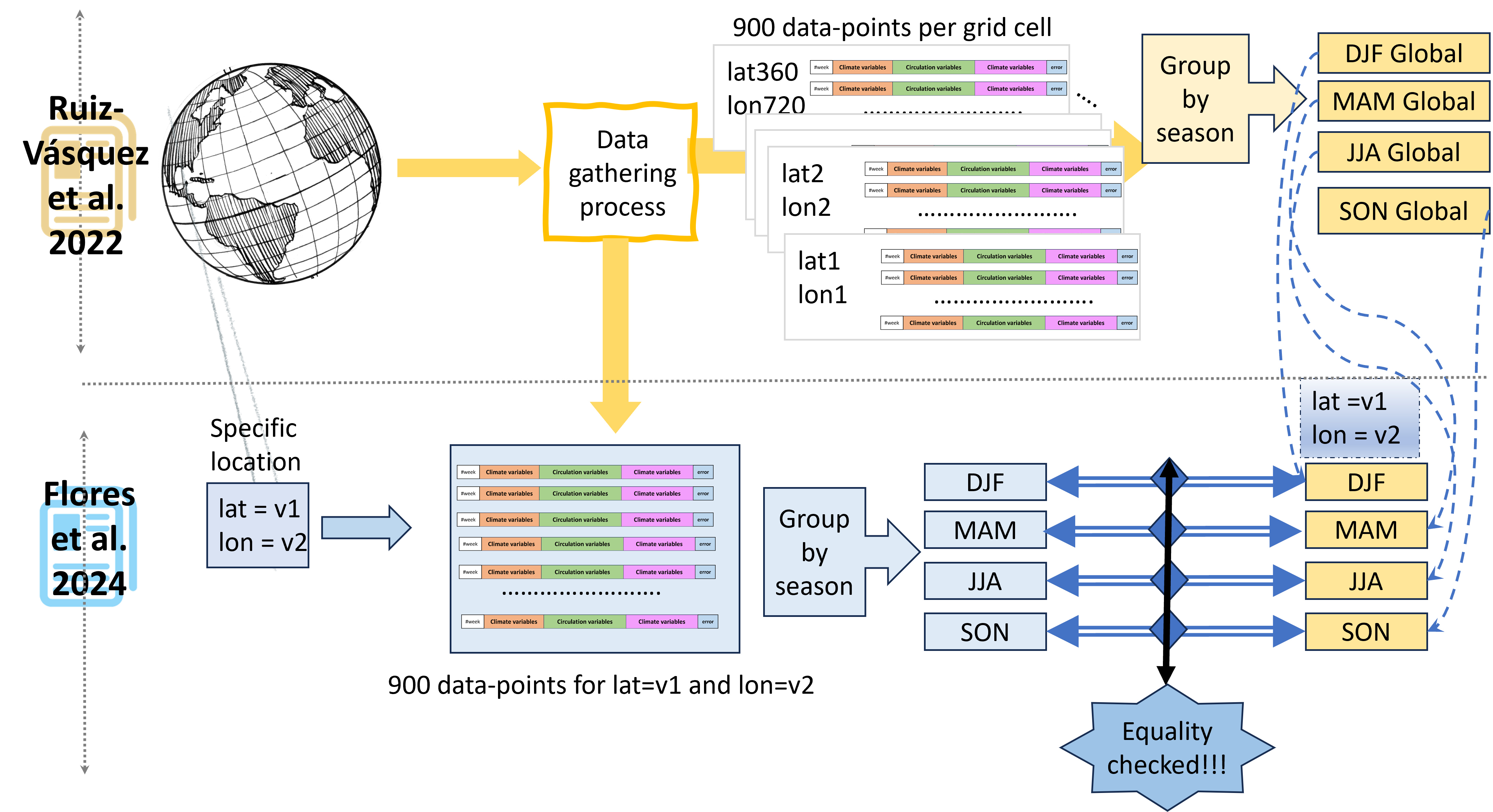}
     \caption{This figure shows the parallel procedure to get the values focusing on a particular location starting from the original research work.}
    \label{fig:data-frame-checking}
\end{figure}

To conduct the research presented in this paper, the first step was to re-run all the scripts coded by Melissa Ruiz-V\'azquez for the experimentation shown in \cite{ruiz2022exploring}. Contributing with a more Machine Learning perspective, co-author and visiting researcher in Jena,  M. Julia Flores, needed a better understanding of the data source and of all the intermediate computations, together with insight into how the distinct multi-dimensional datasets were transformed. For this collaboration, we decided to perform a local study to make the conclusions clearer and the analysis oriented into particular locations. That also makes the learning process manageable, and for future work, those tools that prove more promising could be integrated into the worldwide grid. Then, both for verification purposes and simplification (faster process and much smaller input files), the experimentation here presented uses the same workflow as in Fig. \ref{fig:data-gathering} but stores it for a single square in the original gridding system. The global system divided the globe into 360 longitudes, and latitudes were divided into 720 elements. In the current experimentation, we will just use one square but look for meaningful locations. The data verification, that is, how the two processes have been checked to provide identical data, is depicted in Fig. \ref{fig:data-frame-checking}.  The code to construct a data-frame structure for a point was distinct than the original one, being the sources the same as in Fig. \ref{fig:data-gathering}. Also, per location, we save the yearly data, but in the global process, the information is separated into files by season\footnote{December–January–February (DJF), March–April–May (MAM), June–July–August (JJA), and September–October–November (SON)}. Finally, for a specific location, we extract the data frames and check that the values for all the features and targets are identical. Notice that the top part of this graph describes the work done in \cite{ruiz2022exploring}, while the bottom part corresponds to the work performed to carry out the study of this paper.

Table \ref{tab:chosen_locations} shows the locations we have finally considered. Notice that the location name is an approximation. Using the original paper's results, five locations were spotted using the Nominating\footnote{\url{https://nominatim.org/}} API. Every location has a descriptive name, the associated latitude (lat) and longitude (long) coordinates. Then, we include the exact latitude and longitude values in the dataset, stored in multidimensional \textsf{.nc} files. Finally, we report the indexes i\textsubscript{lat} and i{\textsubscript{lon}}, which are internal values that serve for checking the data are correctly stored and extracted from the global dataset (see right part in Fig. \ref{fig:data-frame-checking}).

\begin{table}[h!]
    \centering
    \begin{tabular}{|l|r|c|c||c|c||c|c|}\hline
    Name & Country & lat & lon &  lat\textsubscript{nc}& lon\textsubscript{nc} & i\textsubscript{lat} & i{\textsubscript{lon}} \\\hline       
       
Lewiston &  USA &  46.42 &  -117.02 & 46.25 & -117.25 & \texttt{87} &  \texttt{125} \\ \hline

Lincoln &  USA &  40.81 &  -96.71 & 40.75 & 96.75 & \texttt{98} &  \texttt{166} \\ \hline

Jena &  Germany &  50.93 &  11.59 &  50.75 & 11.75 & \texttt{78} &  \texttt{383} \\ \hline

Jiangxi &  China &  28.00 &  116.00 & 28.25 & 116.25 & \texttt{123} &  \texttt{592} \\ \hline

Canberra &  Australia &  -35.30 &  149.10 &  -35.25 & 149.25 & \texttt{250} &  \texttt{658} \\ \hline
    \end{tabular}\vspace*{0.2cm}
    \caption{Main info about the chosen locations}
    \label{tab:chosen_locations}
\end{table}

\subsection{Exploratory Analysis}
In the original paper, Spearman correlation was computed per season and per location to generate rankings of variables; the main results were analyzing those variables in the first positions. For a better generalization, they also used the three families of variables (Land surface, Circulation and Climate). In this previous study, only significant correlations were finally used. To compute that significance, the Benjamini–Hochberg procedure \cite{benjamini1995controlling} was applied to ensure control of the false discovery rate. The limitation of this approach is that in many locations, a great number of the variables did not research the significance threshold, and their values were lost for analysis purposes. In this new approach, we will use all the correlation values without applying a post-process to the specific values. In our case, the motivation is that the tools we will apply for machine learning will be able to detect those spurious relationships. For that purpose, we will generate an aggregate ranking that integrates the information from correlation metrics. Also, we will use a top-k strategy to use the most important ones.

\begin{figure}
    \centering
    \includegraphics[trim={1.35cm 3.25cm 1.75cm 2cm},clip,width=9cm]{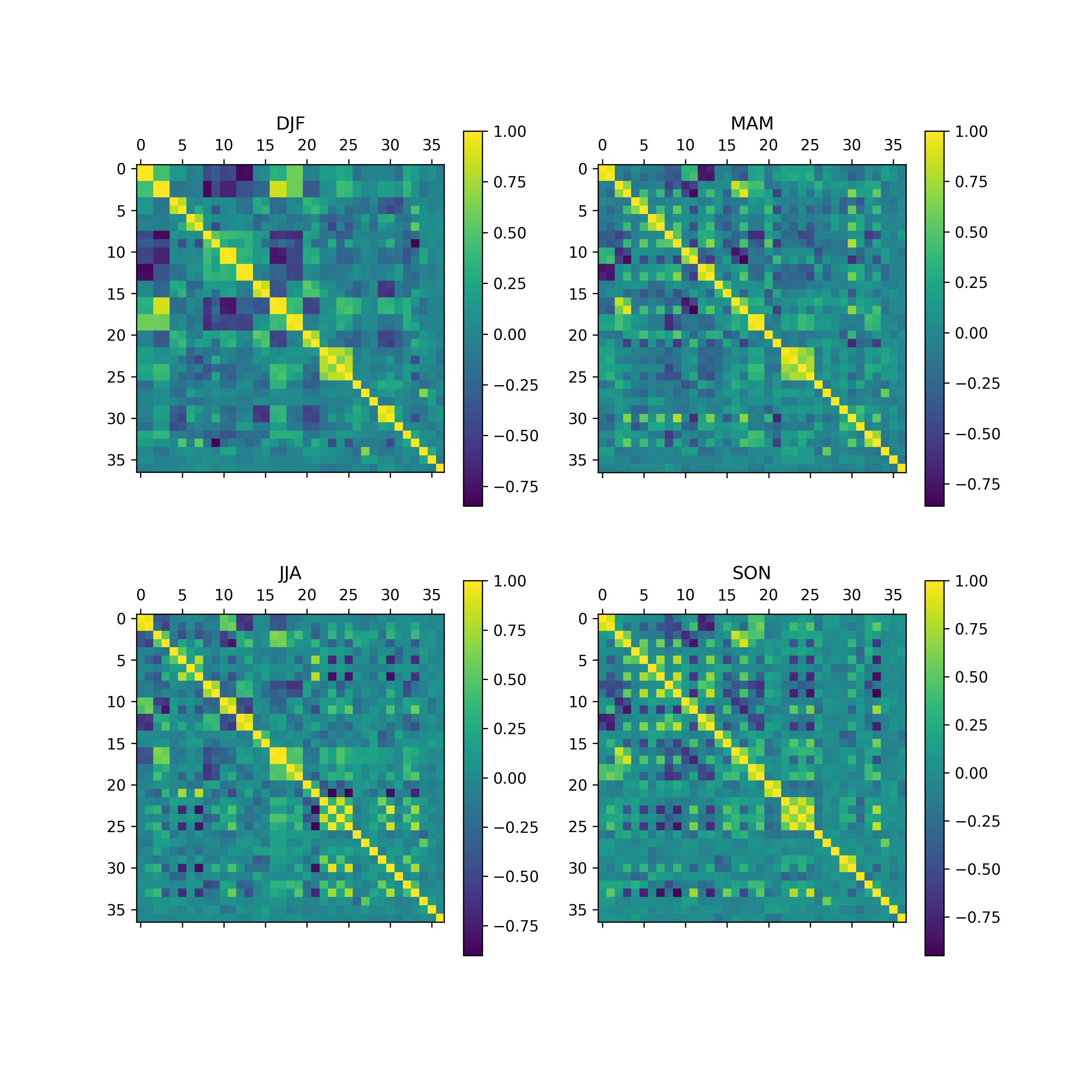}
    \caption{Spearman Correlation for Lewiston, United States.}
    \label{fig:spearman-heatmaps-lw}
\end{figure}

With the data frame corresponding to one of our locations, Lewiston, Fig. \ref{fig:spearman-heatmaps-lw} plots the Spearman correlation of every variable with respect to every variable in a heatmap, distinguishing by season. In this heatmap we do not identify the specific variables. Each of them is shown with an index from 0 to 35, and 36 belongs to the error. We can visually identify clusters of variables, as the anomalies and absolute values of the same features are typically together. We can also be how the diagonal always shows 1.

\begin{figure}
       \centering
    \includegraphics[width=9cm]{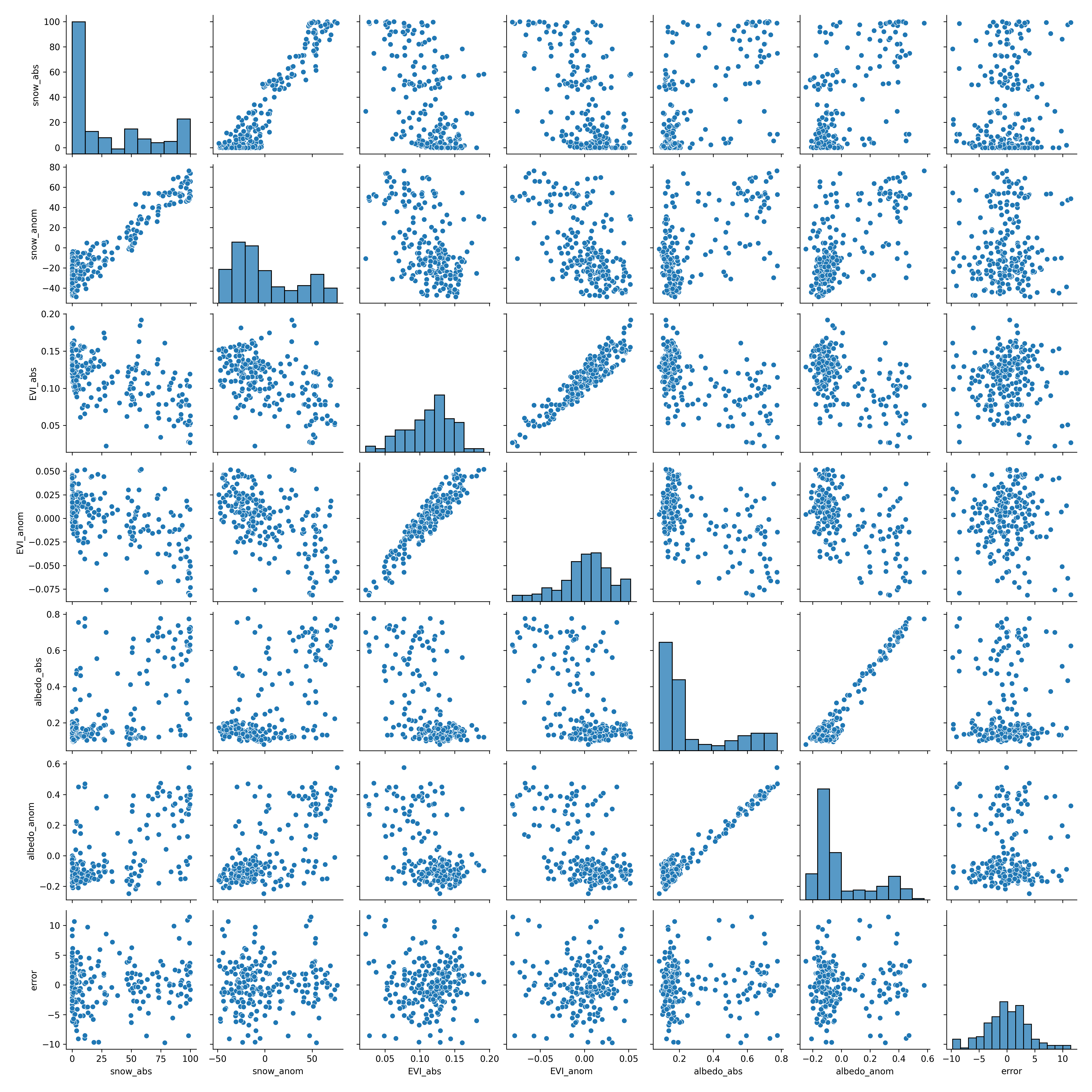}
    \caption{Partial pairplot for Lewiston (USA) in winter (season DJF).}
    \label{fig:lw_winter_pairplot}
\end{figure}

In this dataset, we maintain the whole data for the 900 weeks distributed by season, like DJF: 220, MAM: 238, JJA: 221, and SON: 221. We have performed some exploratory analysis, but it would not be possible to present all the plots in this paper, as there are too many variables and four datasets for location. As a sample, Fig. \ref{fig:lw_winter_pairplot}, we have selected some of the land surface variables and plotted them in pairs. This plot shows the linearity of the two measurements per variable and also that there is not a clear pattern with respect to the target variable error using any of them individually.

\begin{figure}
       \centering
    \includegraphics[width=7cm]{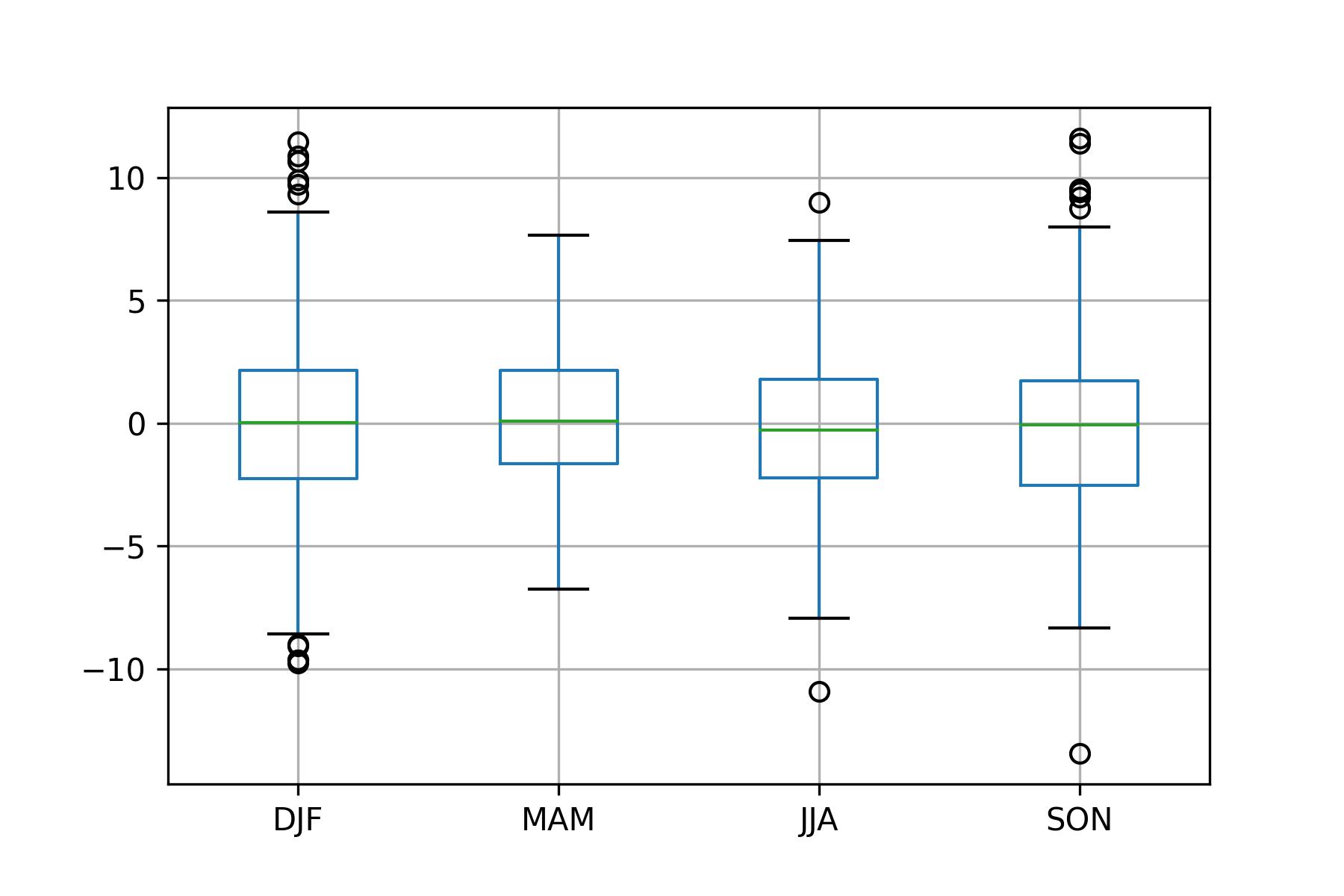}
    \caption{Boxplot for the target variable (error forecast) in Lewiston, by season.}
    \label{fig:error-boxplot-lw-perseason}
\end{figure}

We can see how the error tends to be small, but present some larger values in some cases. This is shown by season in the boxplots at Fig. \ref{fig:error-boxplot-lw-perseason}.

In many exploratory analysis for regression tasks, a Random Forest model is learned with the aim of providing which variables seem more relevant, as commented in subsection \ref{subsec:rf}. As a sample, we have computed the feature importance in winter (DJF) for Lewiston, but also for the spring (MAM), as we can see the first one had a lot of outliers (see Fig. \ref{fig:error-boxplot-lw-perseason}).

\begin{figure*}
    \centering
    \begin{tabular}{cc}
    DJF & MAM \\
    \includegraphics[trim={0.25cm 0.25cm 0.25cm 0.05cm},clip,width=8.25cm]{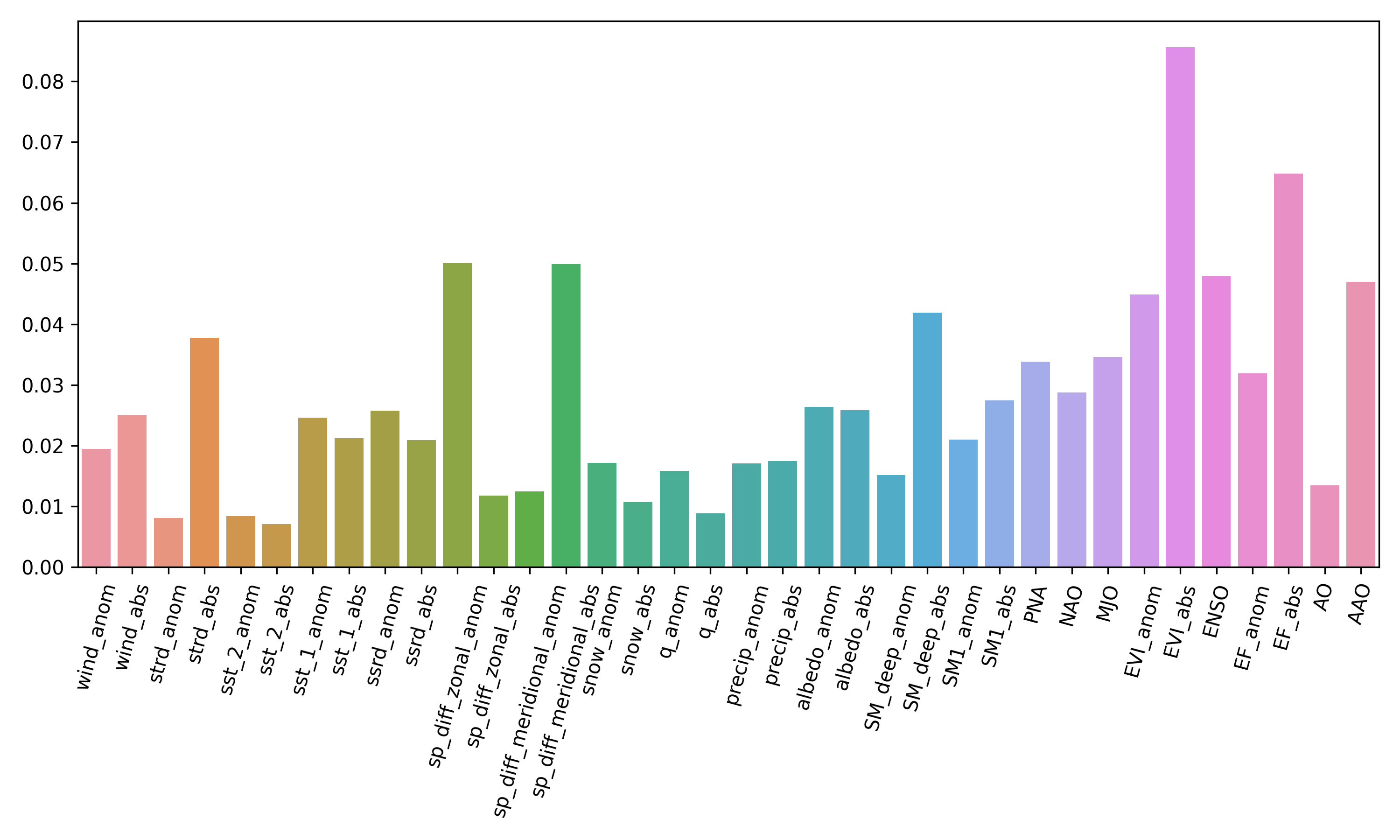} &     \includegraphics[trim={0.25cm 0.25cm 0.25cm 0.05cm},clip,width=8.25cm]{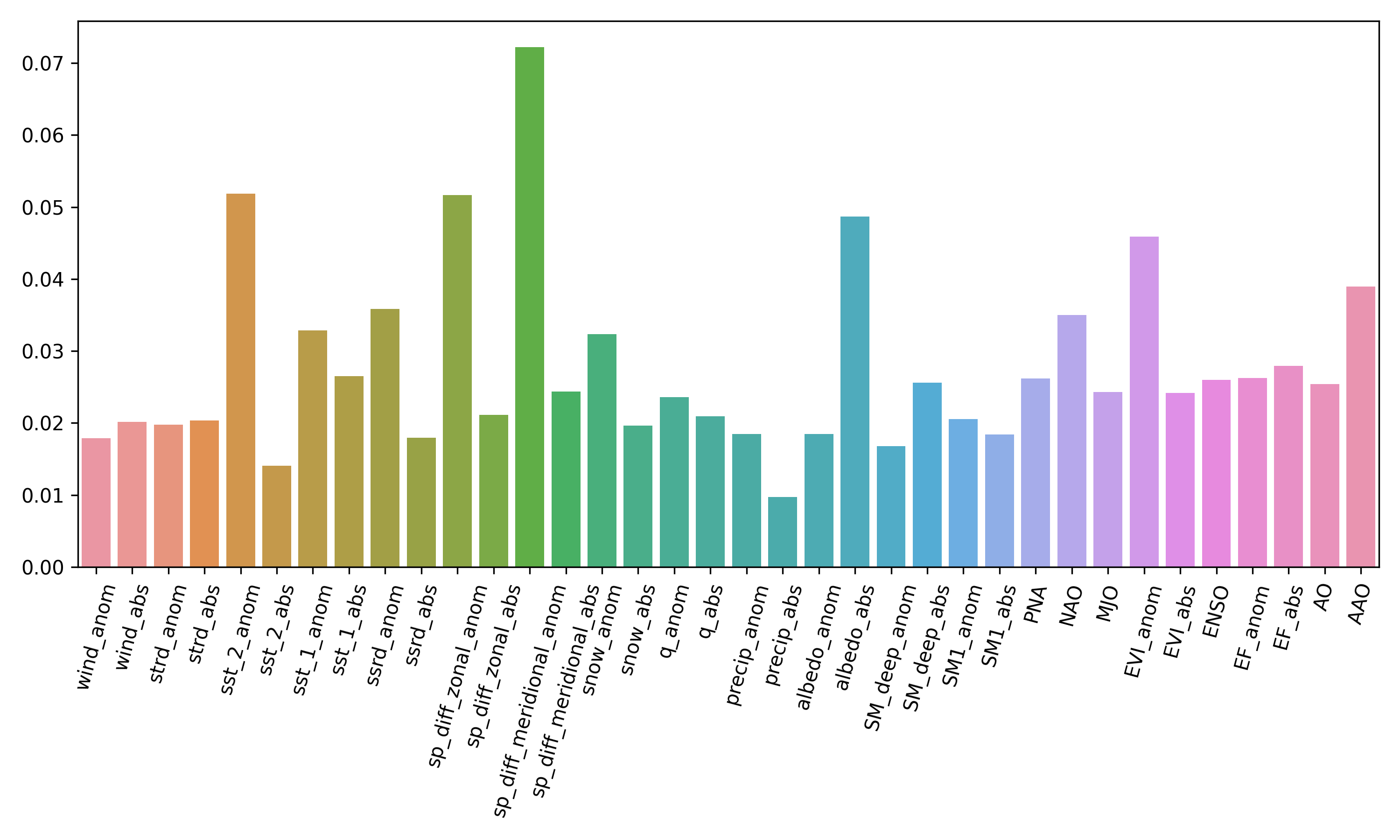}
    \end{tabular}
    \caption{Bar plot that shows the importance by the Random Forest values for two seasons (DJF and MAM) in Lewiston.}
    \label{fig:feature-importance-analysis-rf}
\end{figure*}

\begin{figure}
    \centering
    \includegraphics[width=6.5cm]{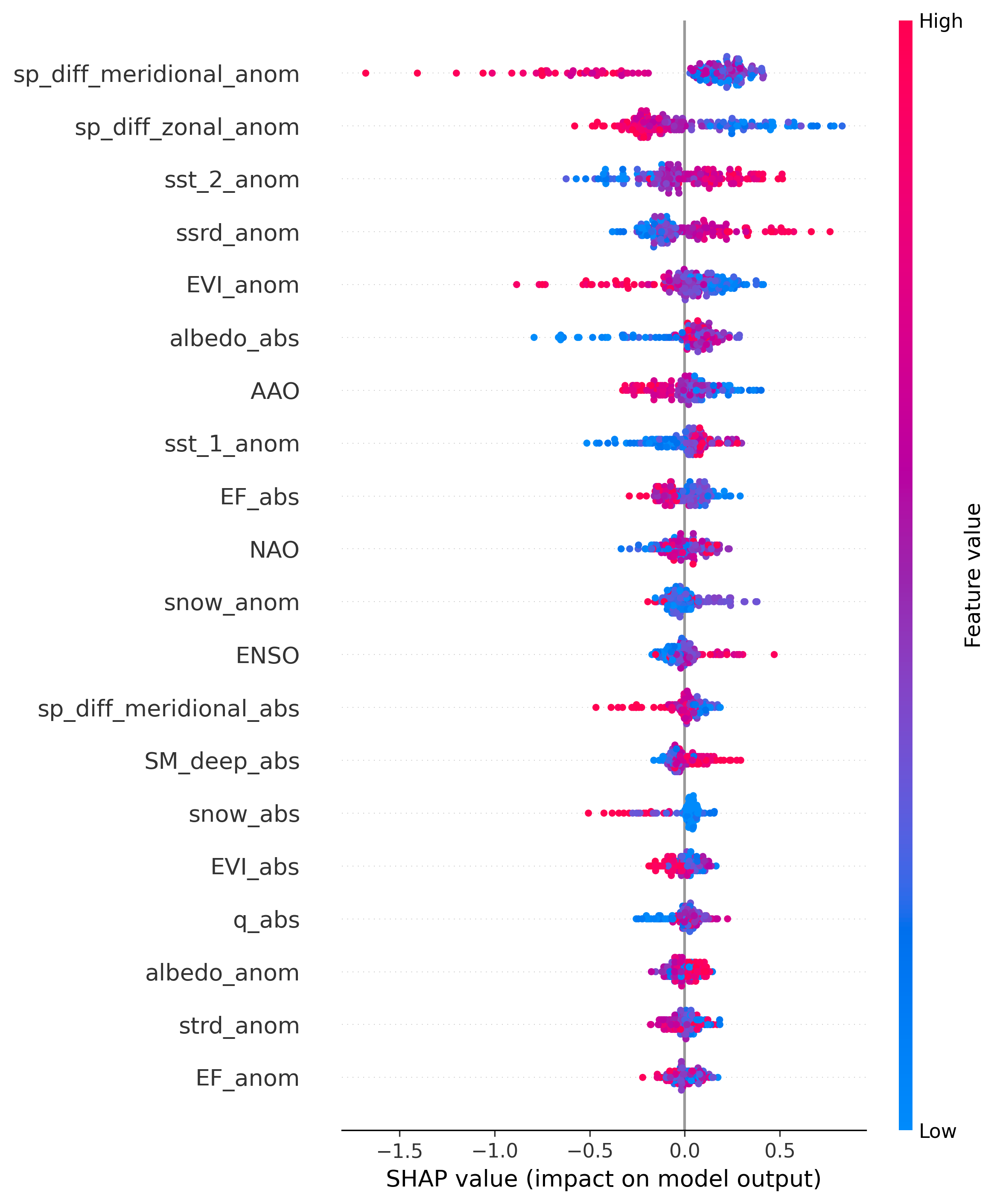}  
    \caption{SHAP summary plot, for Lewiston, MAM.}
    \label{fig:shap-summary-plot-sample}
\end{figure}

\begin{figure}
    \centering
    \begin{tabular}{c}
    SHAP \\
\includegraphics[trim={1.4cm 0 1.4cm 0},clip,width=8.25cm]{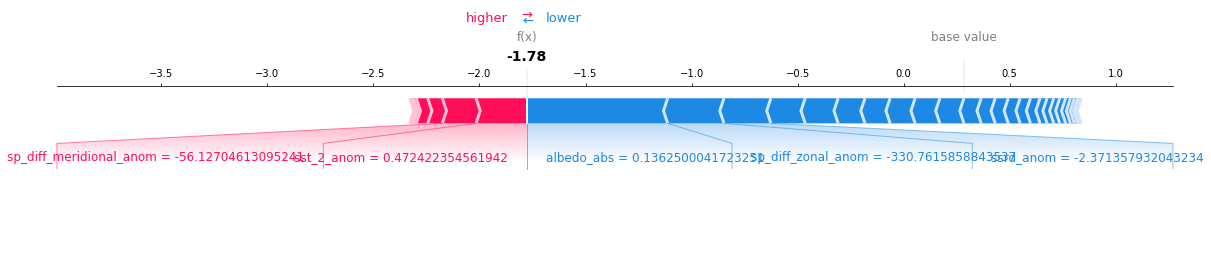} \\ \hline
$~$\\
   LIME \\
\includegraphics[trim={0.8cm 0 0.8cm 0},clip,width=7.1cm]{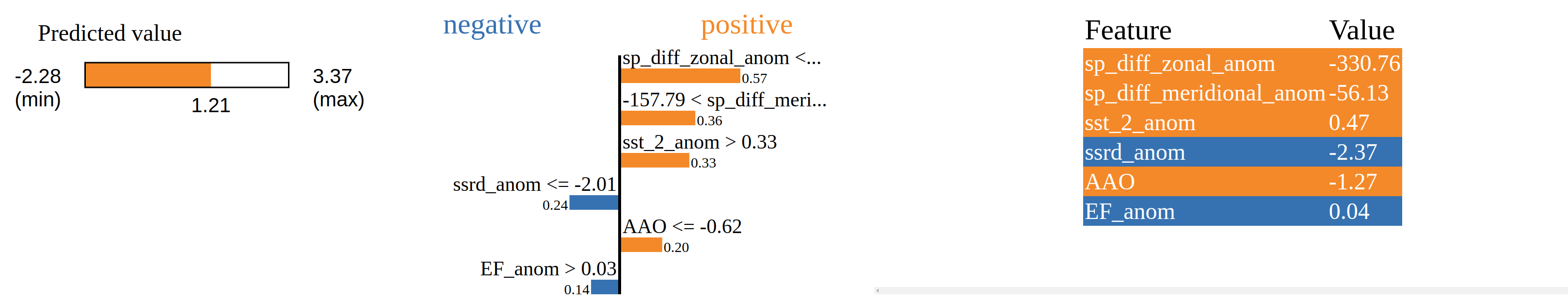}
\end{tabular}
    \caption{Explainability for one random instance both in SHAP (up) and LIME (down).}
    \label{fig:shap-lime-case-18}
\end{figure}

Because we have a model able to predict, in this analysis we could use the explainability techniques described in \ref{subsec:explainability}. For those, we should use, a model. As a sample, using the random forest learn for the 'MAM' season, and keeping 30\%  of the test to train in Lewiston, Fig. \ref{fig:shap-summary-plot-sample} shows the summary plot for the SHAP explainer. LIME is tailored for local explanations, that is, given a prediction, indicate which variables played an important role. This possibility is also possible with SHAP. To show a possible use, we randomly picked an instance and used both tools, and plotted the results in Fig. \ref{fig:shap-lime-case-18}. It is very illustrative for the use but will be interesting for analysing particular predictions.

This brief exploratory analysis can give an idea of the kind of data we are working with. However, this only refers to a particular cell in the original 720x360 grid created in \cite{ruiz2022exploring}, as the final goal is to make a global study that accounts for more information. All the plots provided, as well as more extended analysis and pre-processing, would be really interesting when the expert wants to focus on a particular location. But this is not the scope of the current work. Our current study is about searching for a more informed methodology to analyze the influence of the earth variable on the error forecast with a common framework or methodology that could be applied to all the squares. This is a preliminary work that could be extended in many directions, some of them discussed in section \ref{sec:conclusions-further-work}. The five locations described in subsection \ref{subsec:chosen_locations} will serve for our experimentation that could help us evaluate, with an objective metric, the performance of the proposed method. 

\subsection{Experimentation setup}

This work tries to determine if the aggregating method of rankings provides good results. The best way to quantify the performance is by model evaluation. We have chosen Random Forest given its great performance, versatility, and innate ability to provide a selection of the chosen features. The explainability tools are not integrated within the experimentation, as there is no clear and systematic way to validate them.  They could be, however, of great help when trying to visualize unexpected results locally. 

The experimentation will be focused on comparing these three scenarios:

\begin{enumerate}
    \item The performance of the aggregated rank. We will use the top $k$ selected variables and learn a model, in which we can measure performance in the test cases.
    \item The performance using the top k features using the feature importance of the Random Forest model.
    \item As a reference, we will also report the error when learning using all the variables.
\end{enumerate}

There is a key parameter in the experimentation to be reported: how many elements within the rank we are going to use. We will call this parameter $k$ and that set of relevant features or variables will be denoted as top-$k$. It would seem that lower values of $k$ will give higher errors, if we use all the variables, we will be in the base case, that will quite probably tend to overfitting. That is why we tested $k$ in this set of values \{5,10,15,20,25\}. We will report those results that resulting more representative. To compare the experimental results, we report the Root Mean Square Error (rsme). It accounts for the standard deviation of the residuals (prediction errors). This error is commonly used in climatology, forecasting, and regression analysis.

Having an input dataset $X$ and a numeric target variable $y$ is calculated as the square root of the average of the squared differences between the predicted values $\hat{y}$ and the actual values $y$:

\begin{equation}
   rmse= \sqrt{\frac{1}{n} \sum_{i=1}^{n} (y_i - \hat{y}_i)^2}
\label{eq:rmse}\end{equation}

Where:
\begin{itemize}
    \item $n$ is the number of samples in the dataset.
    \item $y_i$ is the actual value of the target variable for the $i$-th sample.
    \item $\hat{y}_i$ is the predicted value of the target variable for the $i$-th sample.
\end{itemize}

The scheme of the global experiments we have launched is described in pseudocode in Alg. \ref{alg:experiment_procedure}. It is important to comment that RF stands for Random Forest (algorithm \ref{alg:RF}). When we use the syntax \textsc{Learn\_Validate(model,data)} we are using a regressor model, the data instances divided in train (80\%) and test (20\%). The model is learned with the training set, and the rsme is reported using the predictions on the test. \textsc{get\_top\_K\_vars} uses the feature importance values and returns those $K$ feature with the highest scores. 

\begin{algorithm}
\caption{Experimentation procedure}\label{alg:experiment_procedure}
\begin{algorithmic}[2]
\Require $K$ (input parameter)
\State \textbf{for every season $s$}
\State \quad \textbf{for every locations $loc$}
\State \quad \quad data $\leftarrow$ \textsc{get\_data(s,loc)}
\State \quad \quad \emph{I. Get top $K$ variables}
\State \quad \quad \quad $r_p$ $\leftarrow$ \textsc{rank\_pearson\_corr}(data)
\State \quad \quad \quad $r_s$ $\leftarrow$ \textsc{rank\_spearman\_corr}(data)
\State \quad \quad \quad $r_k$ $\leftarrow$ \textsc{rank\_kendall\_corr}(data)
\State \quad \quad \quad $borda \leftarrow$ \textsc{Borda($r_p$, $r_s$, $r_k$)}
\State \quad \quad \quad $r_b \leftarrow$ \textsc{rank}($borda$)
\State \quad \quad \quad $sel\_agg \leftarrow$ \textsc{get\_top\_K\_vars($r_b,K$)}
\State \quad \quad \emph{II. Compute rsme}
\State \quad \quad \quad $rf_{base}$ $\leftarrow$ \textsc{Learn\_Validate}(RF,data)
\State \quad \quad \quad $sel_rf$ $\leftarrow$ \textsc{Top\_var($rf_{base}$,K)}
\State \quad \quad \quad $rf_1$ $\leftarrow$ \textsc{Learn\_Validate}(RF,data[$sel\_agg$])
\State \quad \quad \quad $rf_2$ $\leftarrow$ \textsc{Learn\_Validate}(RF,data[$sel\_rf$])
\State \quad \textbf{EndFor}
\State \textbf{EndFor}
\end{algorithmic}\label{alg:exp}
\end{algorithm}

\begin{figure}
    \centering

\begin{tabular}{c}
       \includegraphics[trim={0 5.5cm 0 0},clip,width=8.7cm]{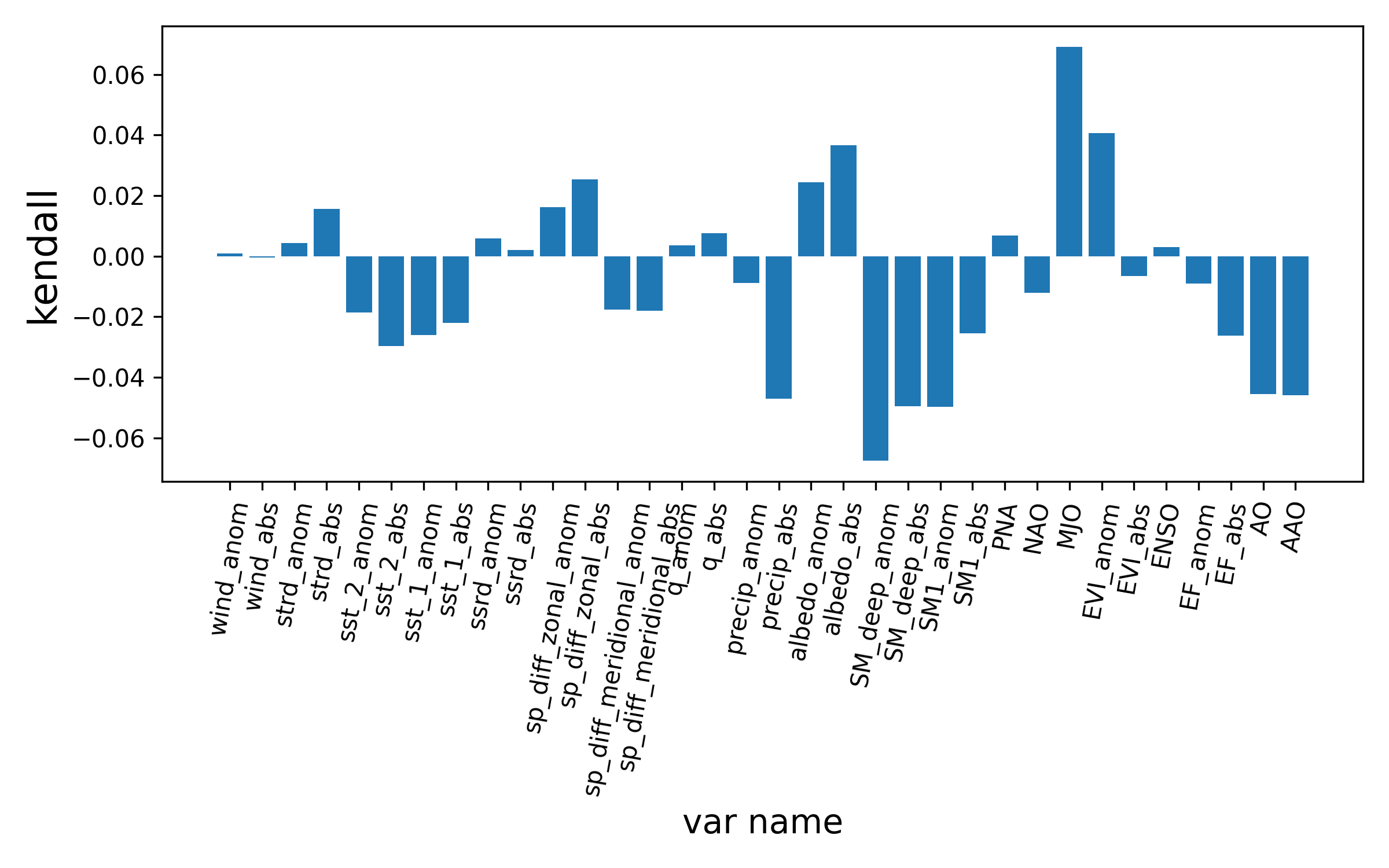}\\
       \includegraphics[trim={0 5.5cm 0 0},clip,width=8.7cm]{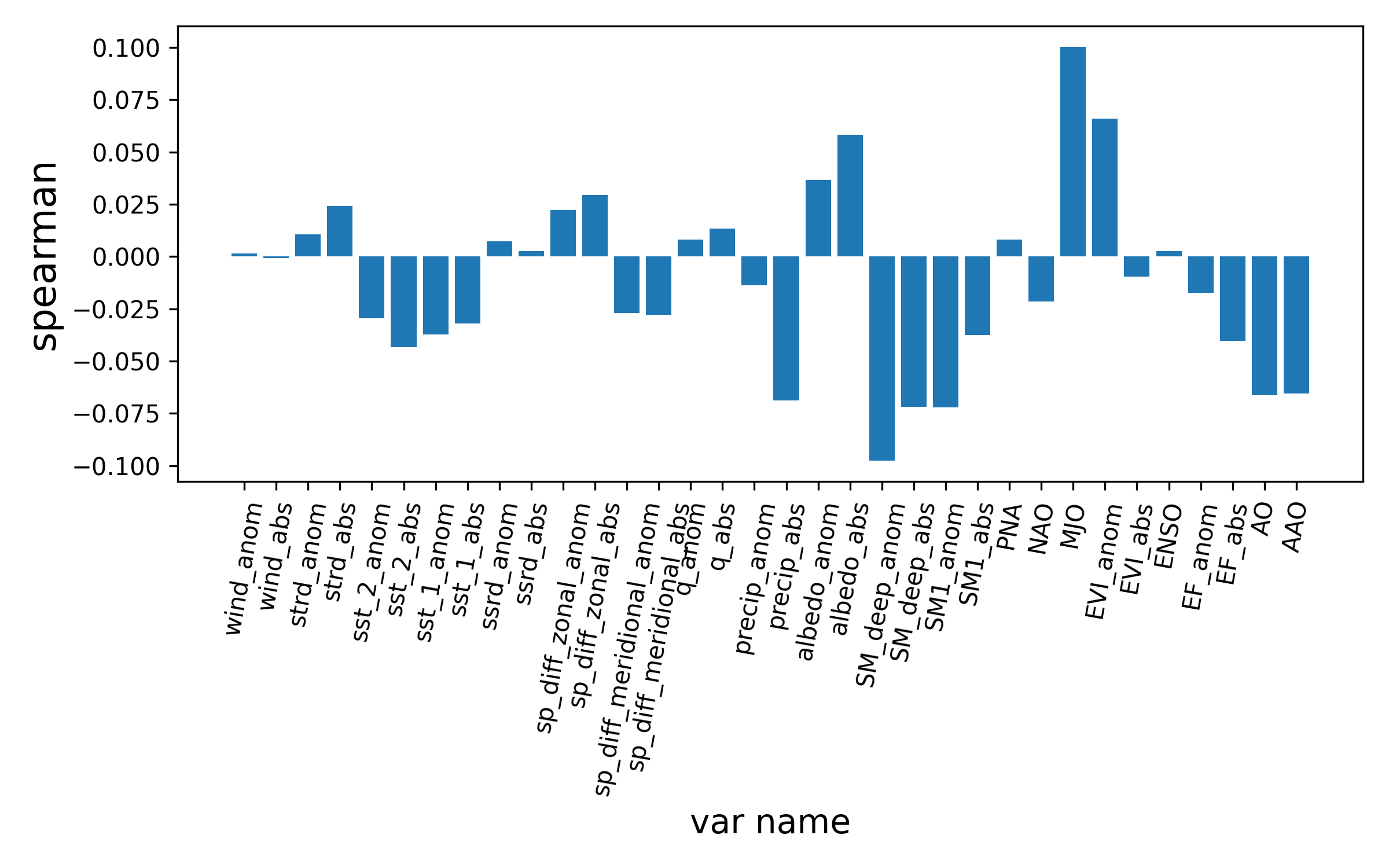}\\
       \includegraphics[width=8.7cm]{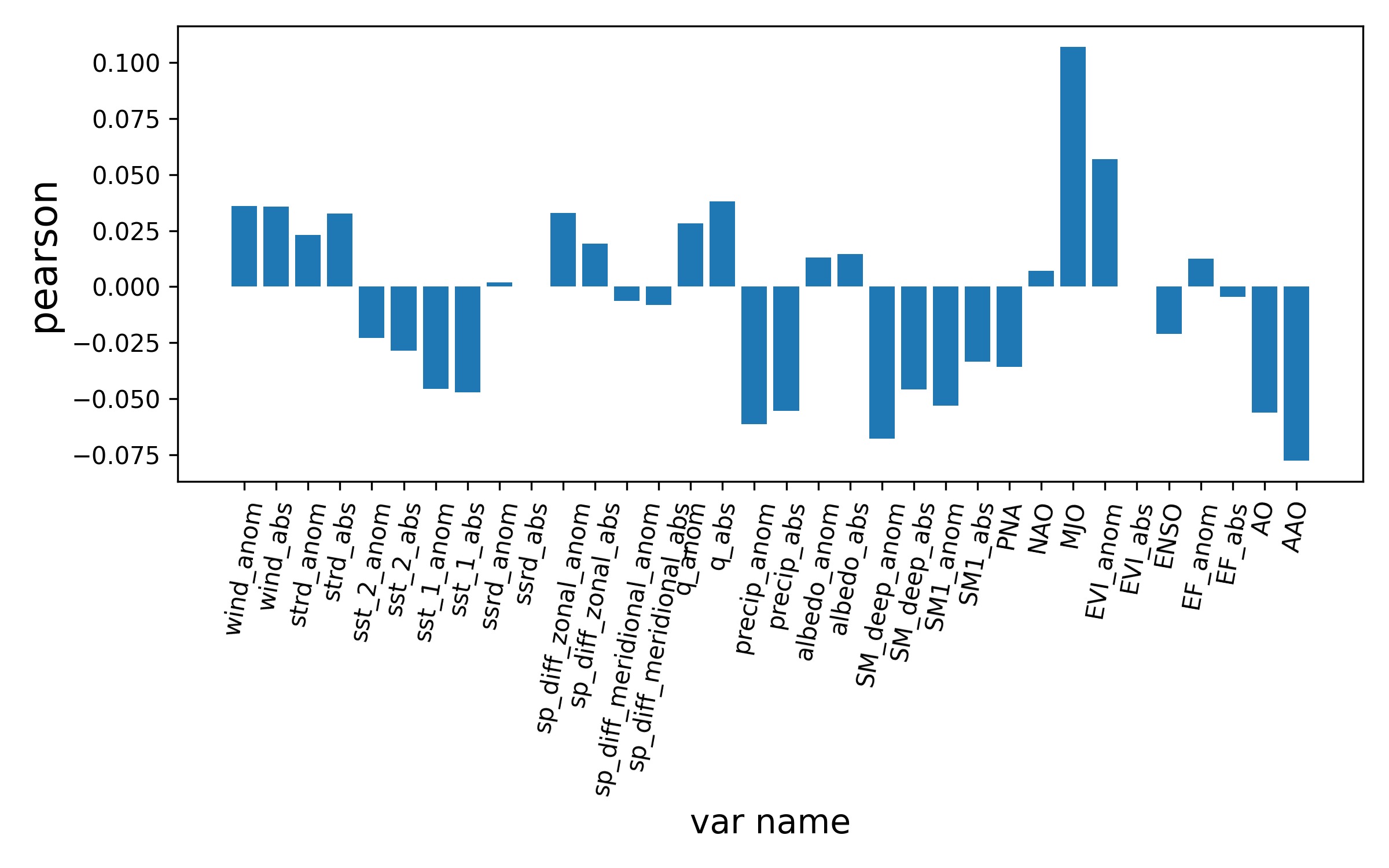}\\
\end{tabular}
    \caption{Bar plot for the three correlation metrics in a single case: Jiangxi in the winter (DJF) season.}
    \label{fig:ranking-jianxi-DJF}
\end{figure}

\begin{figure}
    \centering
    \includegraphics[width=7.5cm]{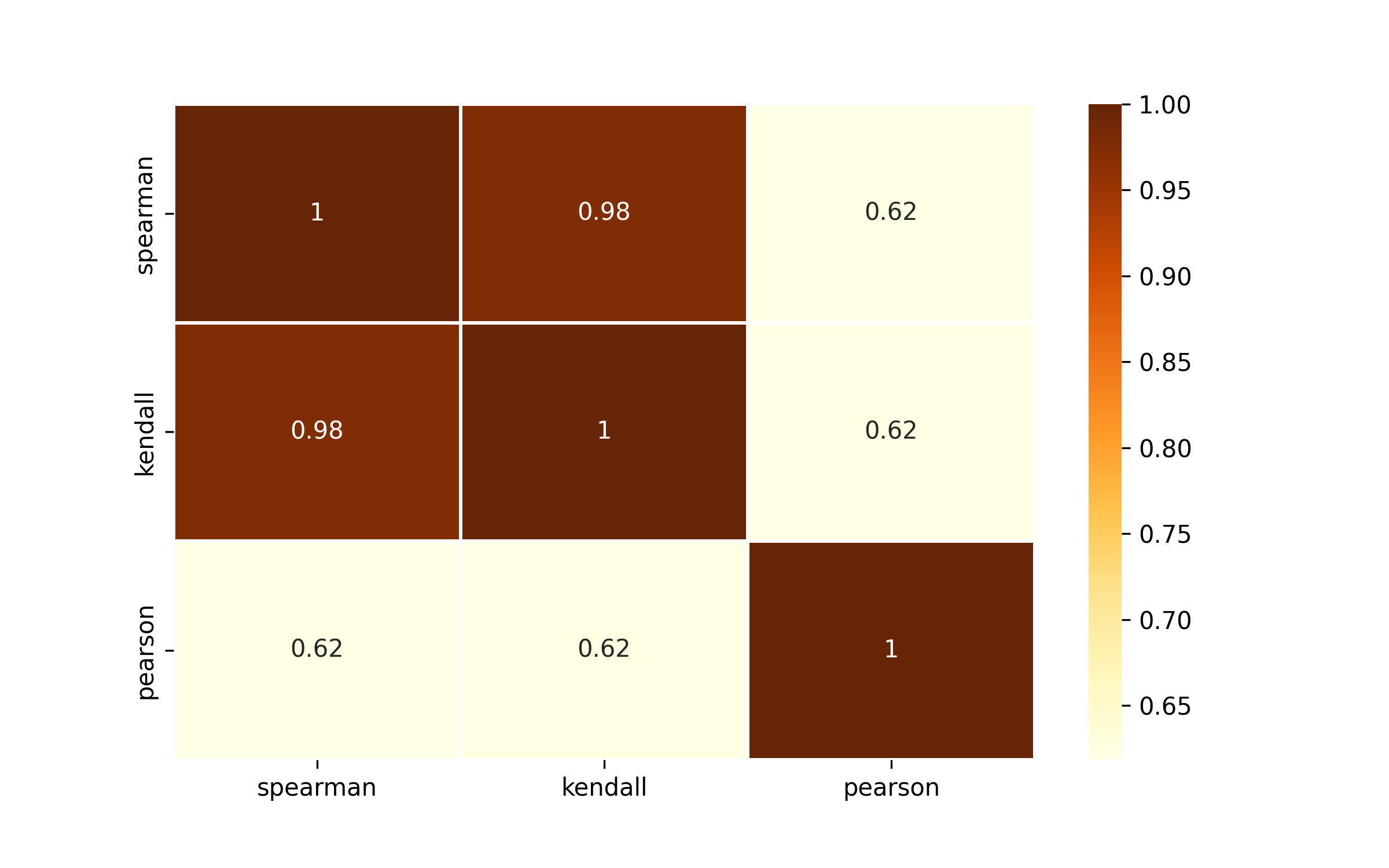}
    \caption{Pairwise correlation of the rankings extracted from the three correlation metrics. Values for a sample case: : Jiangxi in the winter (DJF) season. }
    \label{fig:concordance}
\end{figure}

To illustrate a case for the most important contribution of this paper, Fig. \ref{fig:ranking-jianxi-DJF} shows the bar plots for the three correlation metrics in one sample pair of location and season. In order to measure the concordance between them we compute a concordance value, by the Kendall tau correlation \cite{kendall}, which coincides with the Levenshtein/Wasserstein correlation that \texttt{ranky.corr} method computes. We can see them in the heatmap plot at Fig. \ref{fig:concordance}. In this plot we can see how spearman and Kendall seem closer, while pearson is more distant, having already a correlation clearly higher than 0.5. This pattern repeats in all the cases we have visualized. For the same location in summer season (JJA) this correlation goes to 0.7 and 0.72. In other locations, pearson can be correlated more than 0.8, but the other two metrics are always closer. In this singular case, the average correlation of the original metrics with respect to the aggregated one (by Borda technique) is 0.84.

\section{Analysis of the experiments}

All the experiments were performed using Python code. Apart from the libraries typically used in a Machine Learning (\texttt{scikit-learn}, \texttt{pandas} and \texttt{numpy}), we have used \texttt{xarray} for managing .nc files, \texttt{seaborn} and \texttt{matplotlib} for the plotting, and \texttt{ranky} \cite{pavao2020ranky}
for the ranking computations and visualizations. For plots both \texttt{matplotlib} and \texttt{seaborn} have been applied.

\subsection{Results}

The experimental procedure schemed in Alg. \ref{alg:experiment_procedure}, for every $K$ value yields 4 (number of seasons) $\times$ 5 (number of locations) runs, and every of these runs give us three error values. 

As a starting point summary, table \ref{tab:rf_base_rsme_all} shows the error per season and location without any selection, the one we denoted as $rf_{base}$. Looking at this table, we can see the variability in the error limits, between 2 and 6.25, being season DFJ the most difficult to predict correctly. Besides, we can easily observe how every location behaves differently, which clarifies that the original assumption is true: the earth system variables affect the error forecast. 

\begin{table}[h!]
    \centering
\begin{tabular}{lrrrr}
\toprule
location &     DJF &     MAM &     JJA &     SON \\
\midrule
 Jiangxi & 3.38604 & 2.30072 & 1.85447 & 2.77039 \\
 Lincoln & 5.68711 & 3.84551 & 4.01374 & 3.96976 \\
    Jena & 6.25605 & 4.16224 & 3.20605 & 3.06079 \\
Lewiston & 4.56119 & 2.72669 & 2.98300 & 3.87644 \\
Canberra & 2.03477 & 2.23822 & 2.16543 & 2.46705 \\
\bottomrule
\end{tabular}
\vspace*{0.25cm}
\caption{Errors per location (rows) and season (columns) when using all the variables (Random Forest).}
    \label{tab:rf_base_rsme_all}
\end{table}

\begin{figure*}
    \centering
    \begin{tabular}{cc}
           \includegraphics[trim={0.6cm 0.6cm 0.6cm 0.6cm},clip,width=9cm]{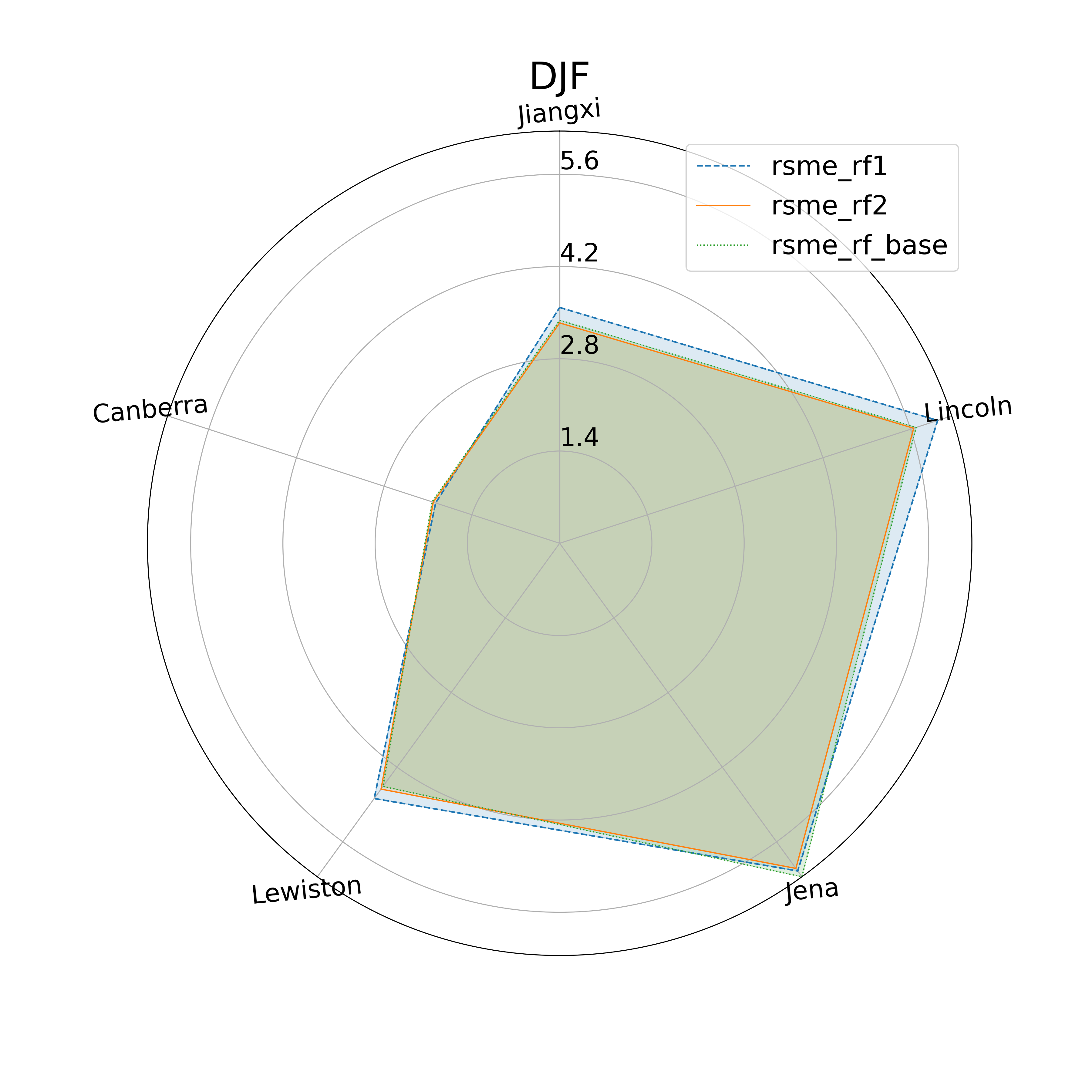} & \includegraphics[trim={0.6cm 0.6cm 0.6cm 0.6cm},clip,width=9cm]{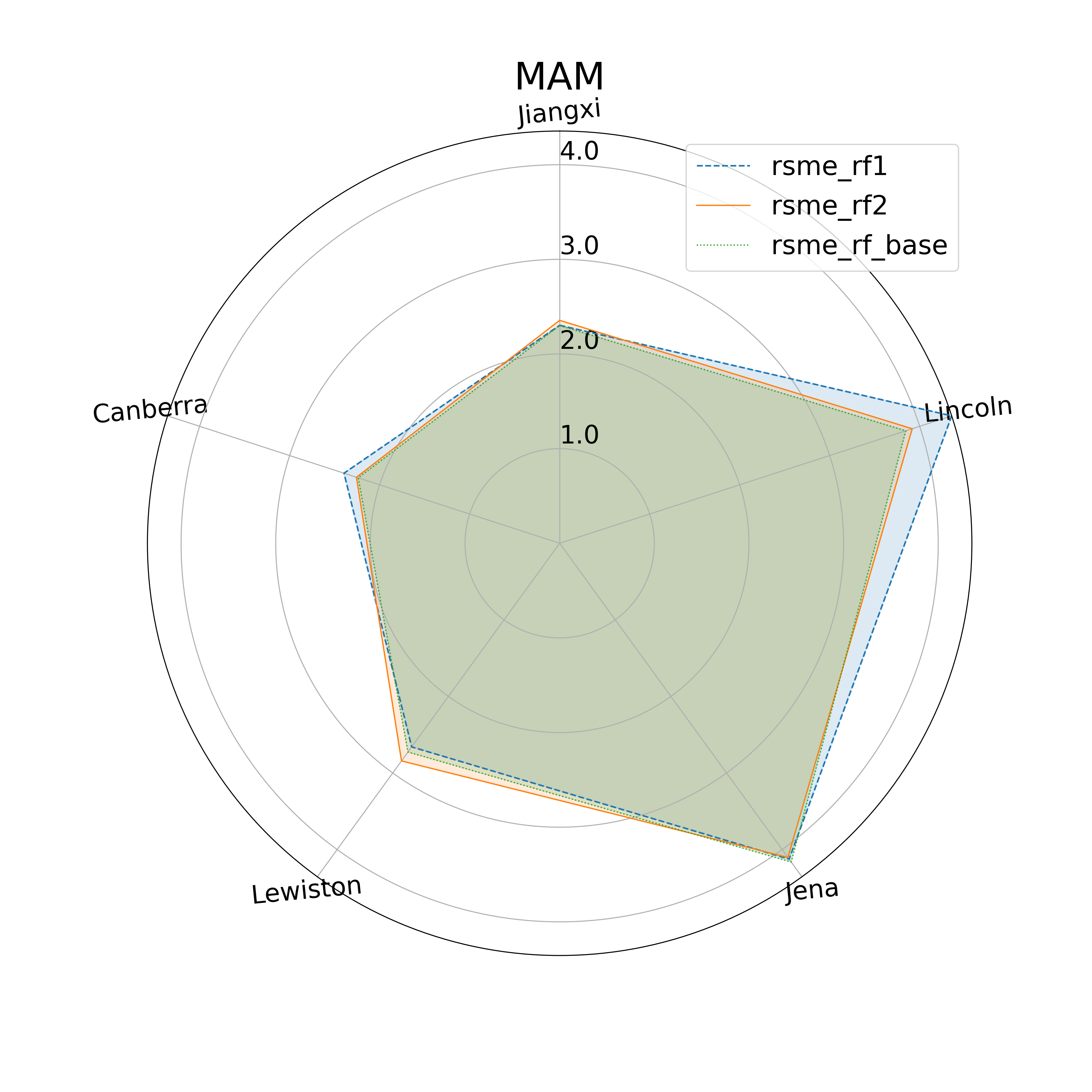}  \\
          \includegraphics[trim={0.6cm 0.6cm 0.6cm 0.6cm},clip,width=9cm]{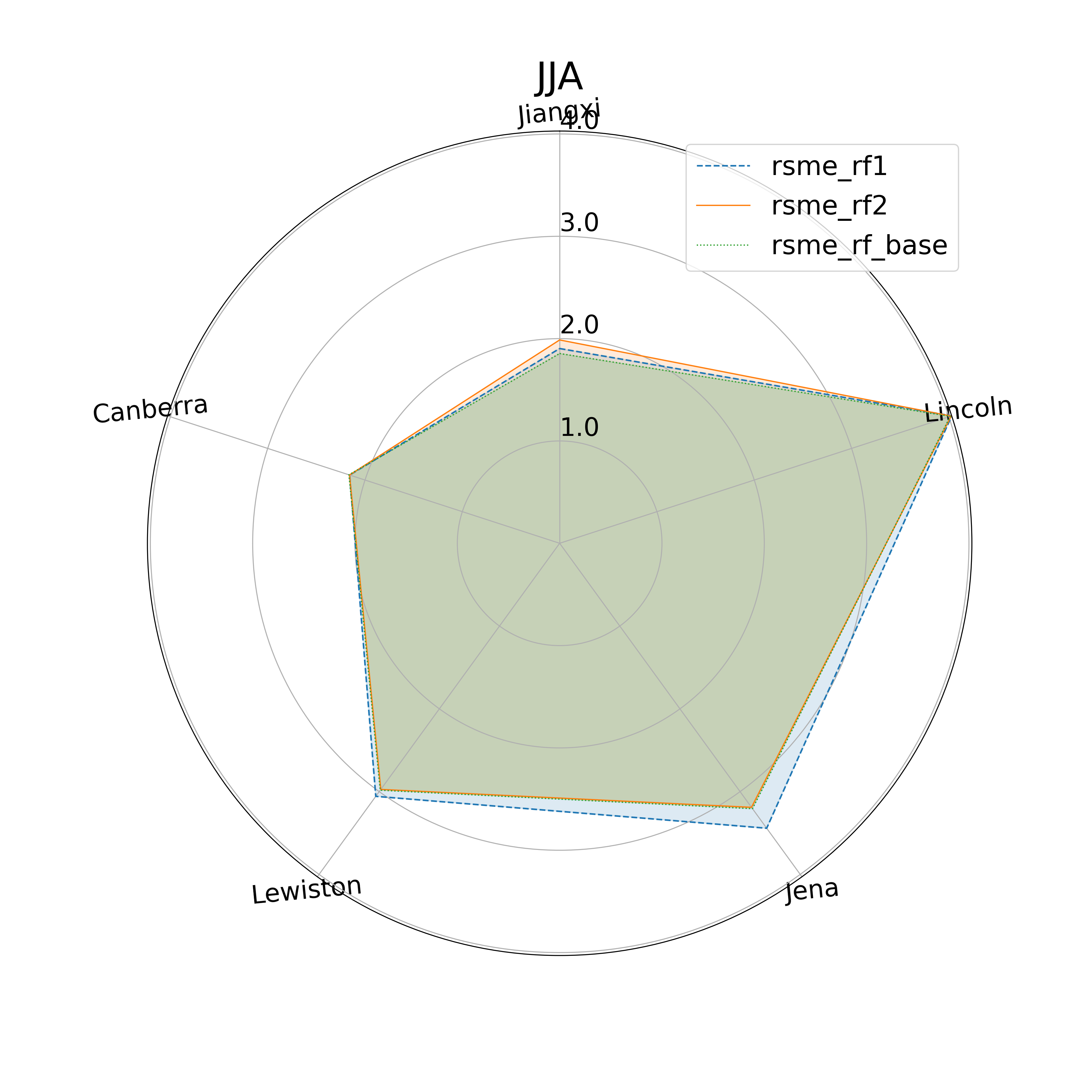} & \includegraphics[trim={0.6cm 0.6cm 0.6cm 0.6cm},clip,width=9cm]{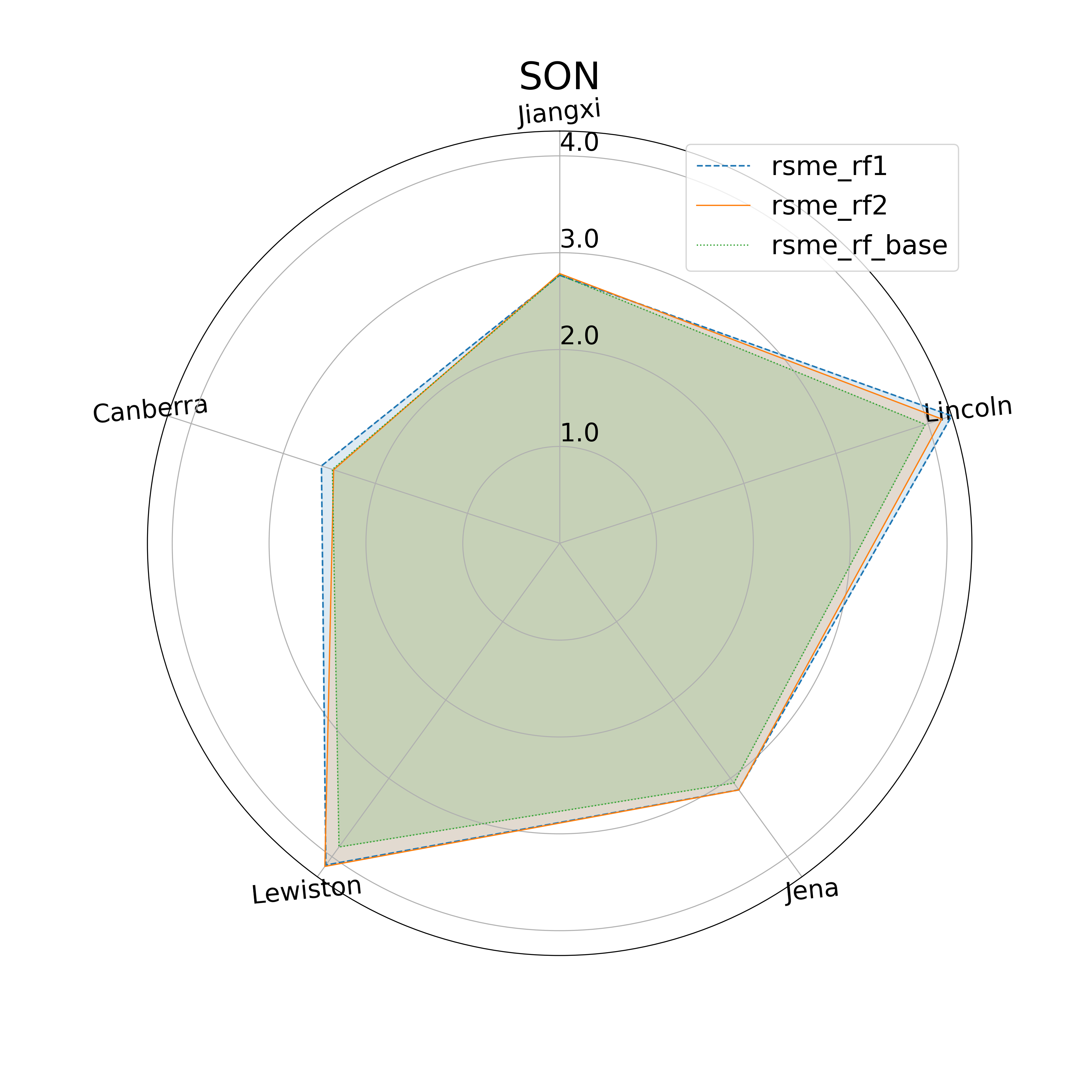}  \\
    \end{tabular}
 
    \caption{Summary radial graphs when $k$ = 15. In the radius we represent every location. Every season is one plot, as the plot title indicates: upper-left is DJF, upper-right is MAM, bottom-left is JJA and bottom-right is SON.}
    \label{fig:radials-k-15}
\end{figure*}

We have designed a compact way of showing the distinct errors for all the seasons and locations. Fig. \ref{fig:radials-k-15} plots this for an intermediate value of $K$ (15). Every radial axis is a location. The filled area is to find if there are clear tendencies. Notice that we are plotting errors, which is why some locations generally have lower values (see Table \ref{tab:rf_base_rsme_all}). Also, if one strategy covers more surface, then the performance is smaller because we intend to minimize error.

\begin{figure*}
    \centering
    \begin{tabular}{cc}
           \includegraphics[trim={0.6cm 0.6cm 0.6cm 0.6cm},clip,width=9cm]{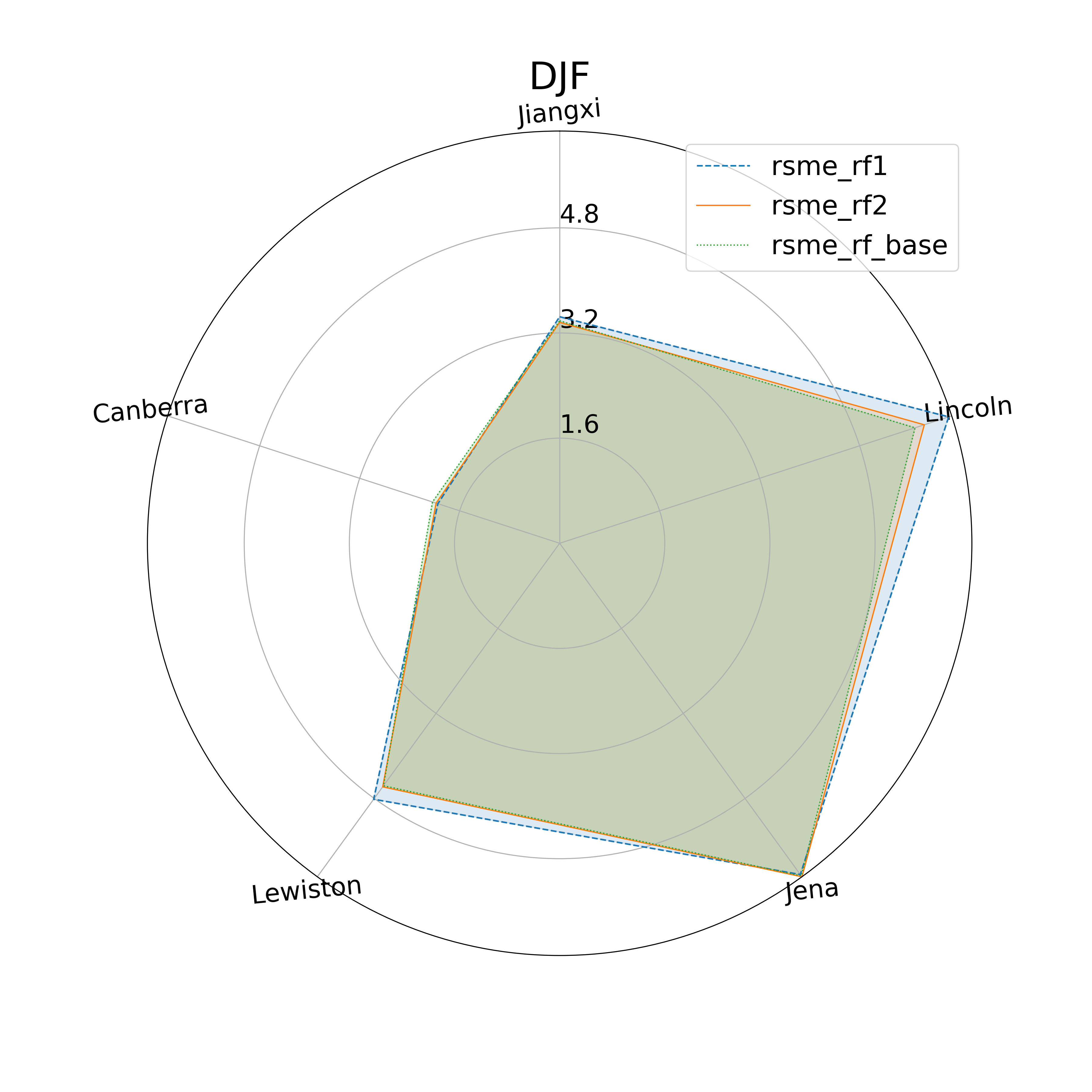} & \includegraphics[trim={0.6cm 0.6cm 0.6cm 0.6cm},clip,width=9cm]{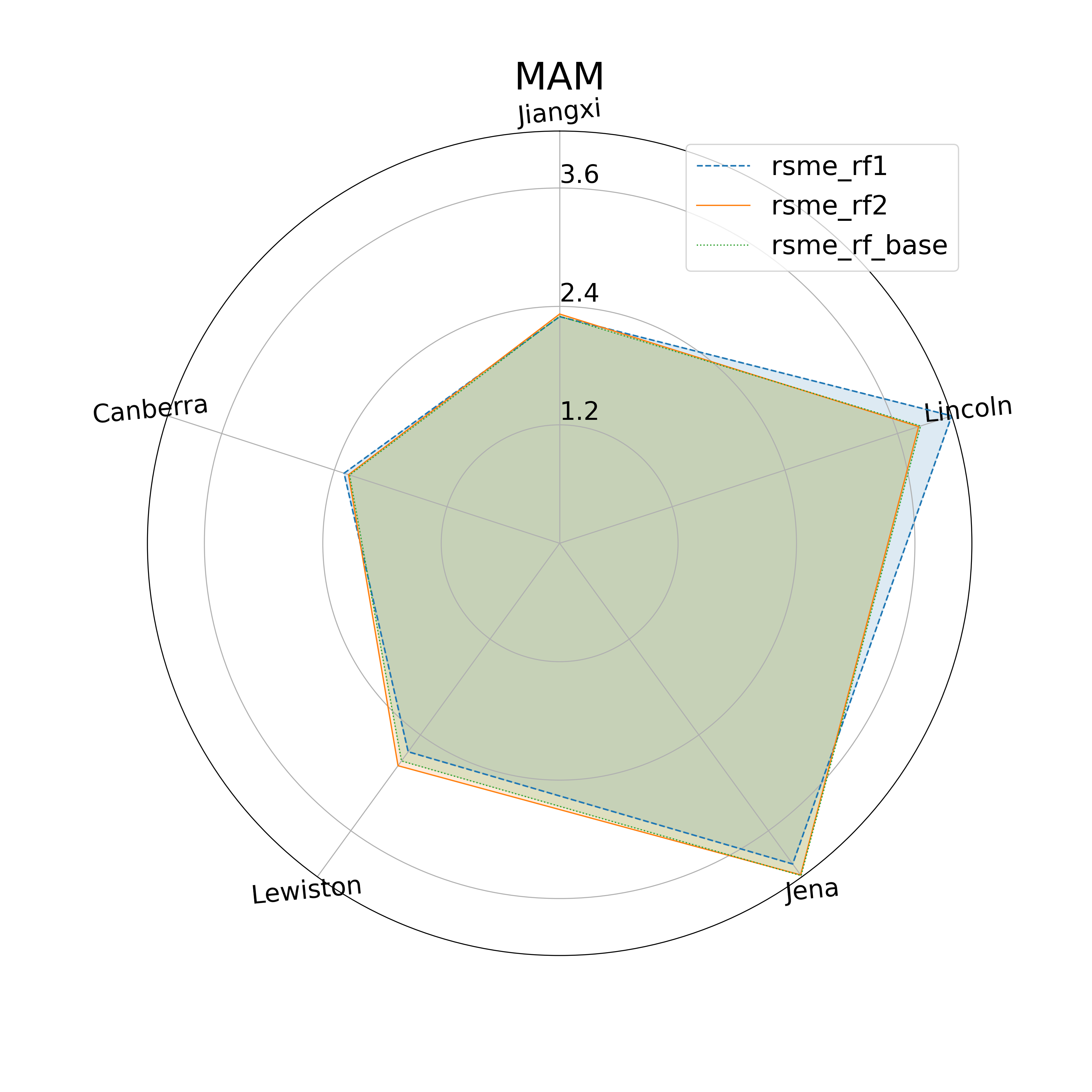}  \\
          \includegraphics[trim={0.6cm 0.6cm 0.6cm 0.6cm},clip,width=9cm]{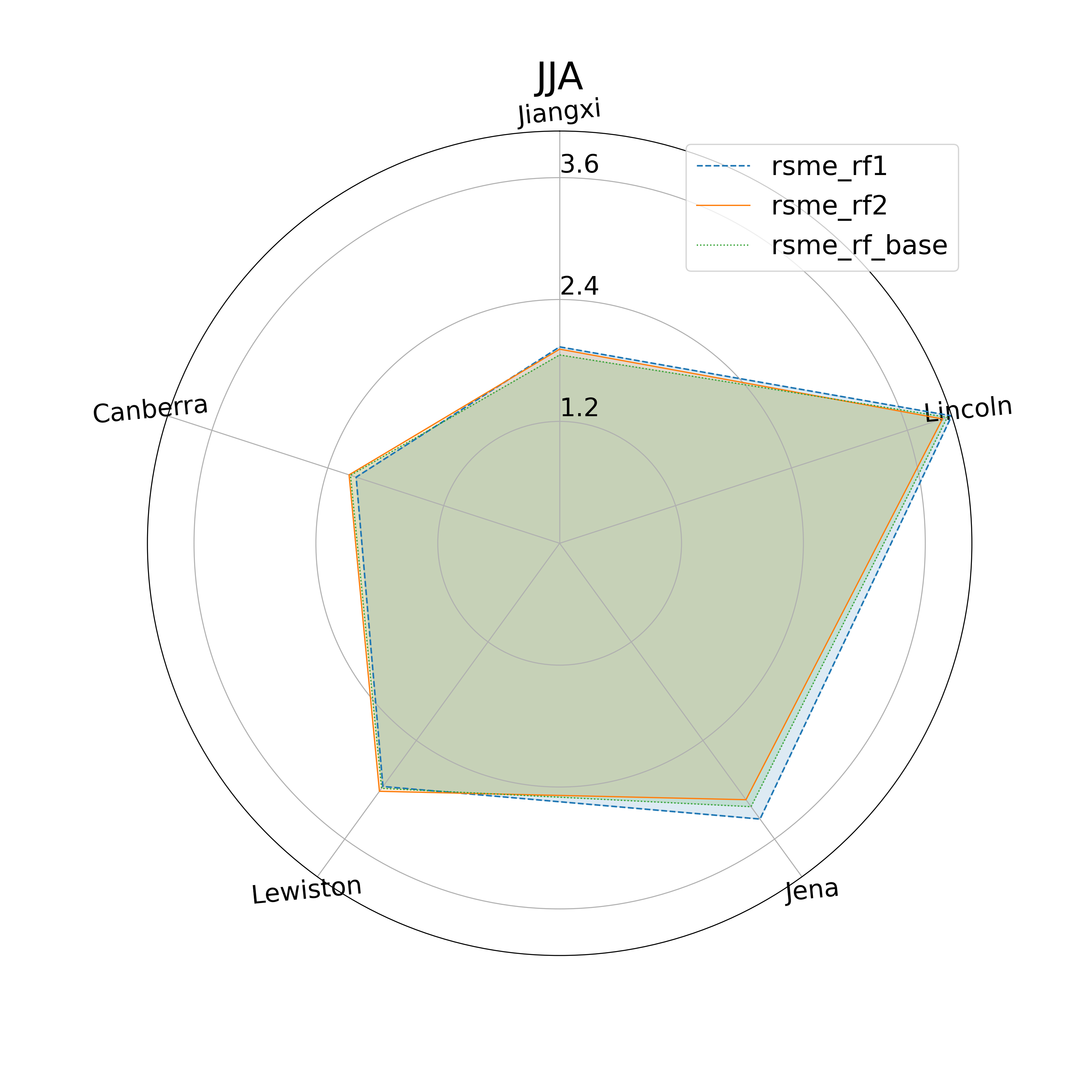} & \includegraphics[trim={0.6cm 0.6cm 0.6cm 0.6cm},clip,width=9cm]{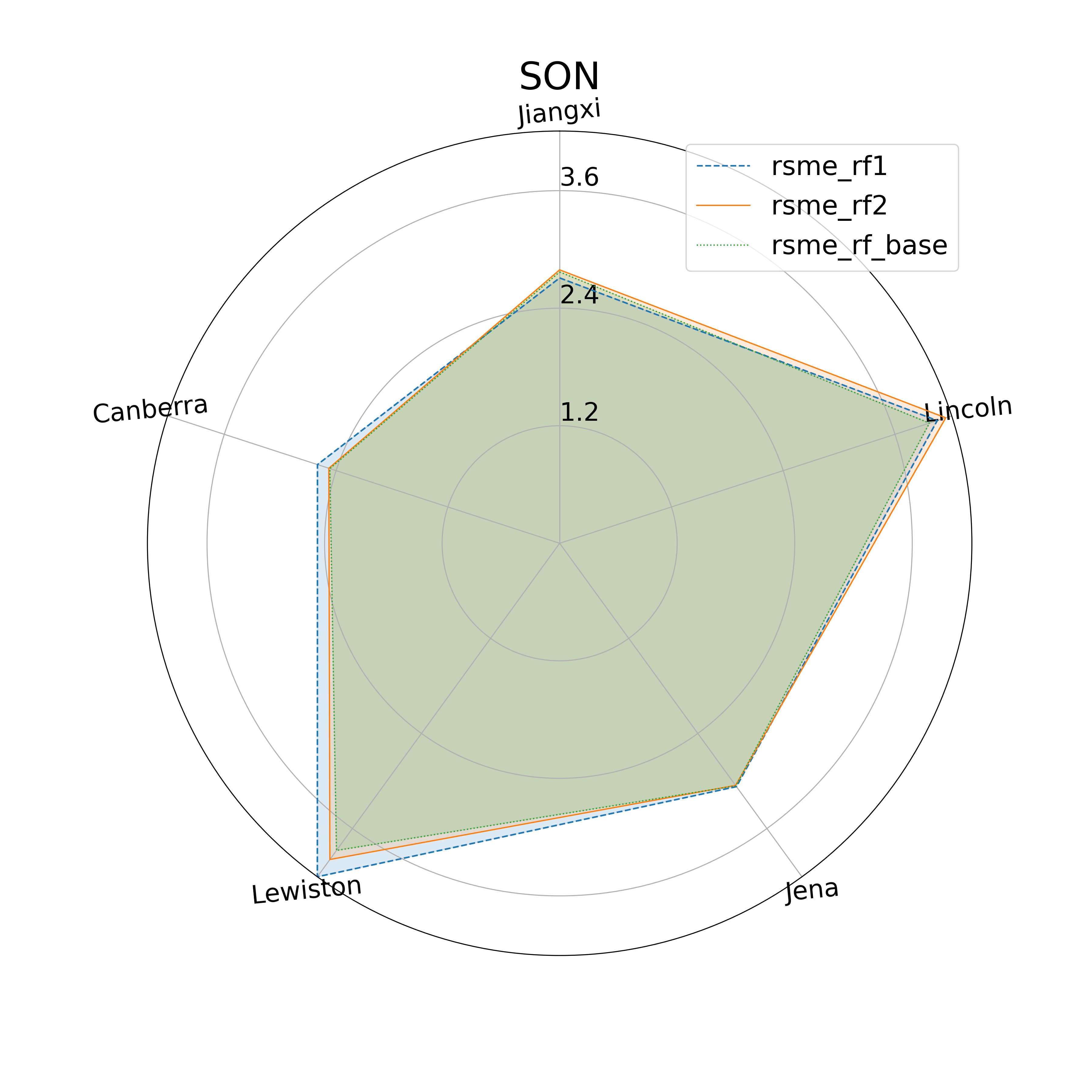}  \\
    \end{tabular}
 
      \caption{Summary radial graphs when $k$ = 20. In the radius we represent every location. Every season is one plot, as the plot title indicates: upper-left is DJF, upper-right is MAM, bottom-left is JJA and bottom-right is SON.}
    \label{fig:radials-k-20}
\end{figure*}

\begin{figure*}
    \centering
    \begin{tabular}{cc}
           \includegraphics[trim={0.6cm 0.6cm 0.6cm 0.6cm},clip,width=9cm]{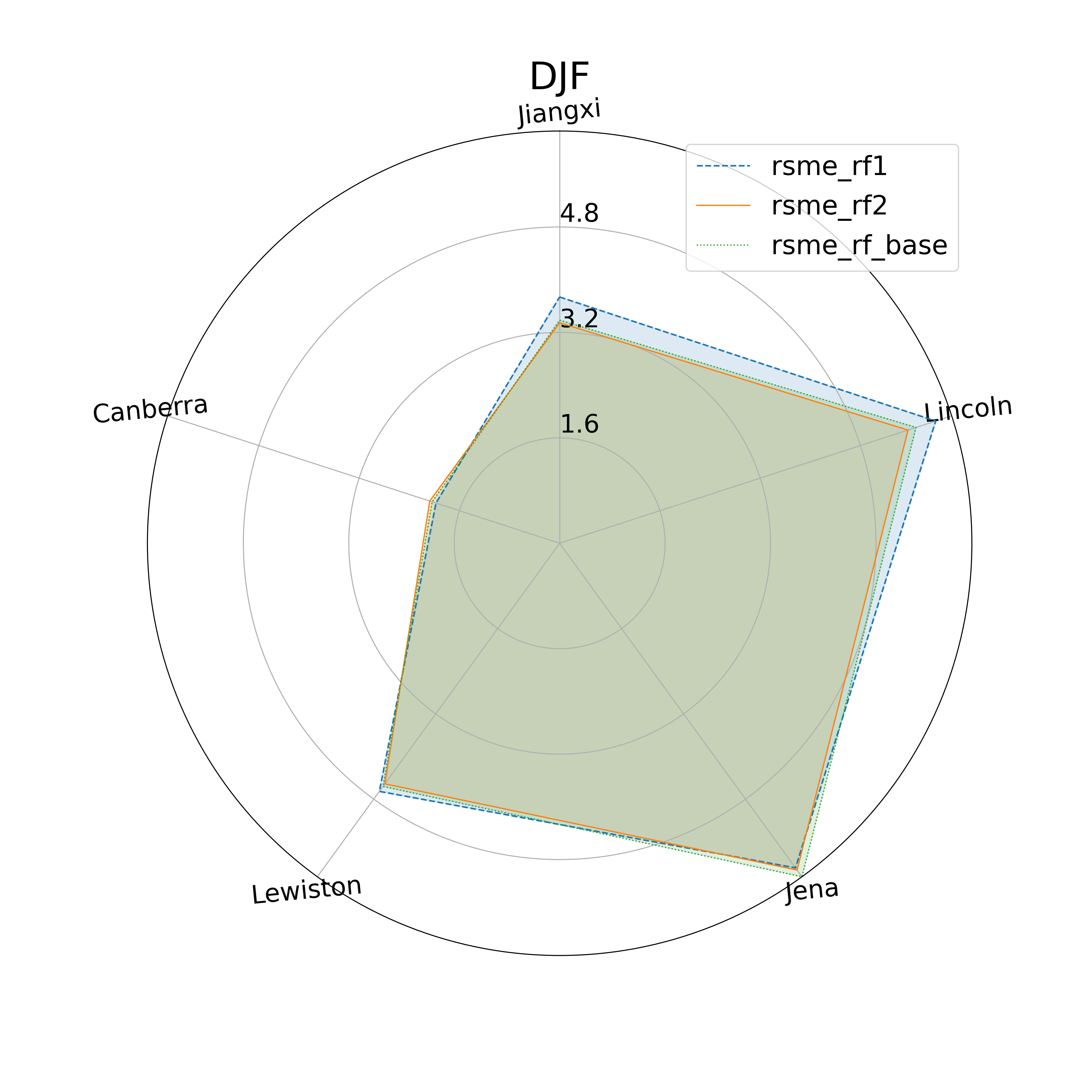} & \includegraphics[trim={0.6cm 0.6cm 0.6cm 0.6cm},clip,width=9cm]{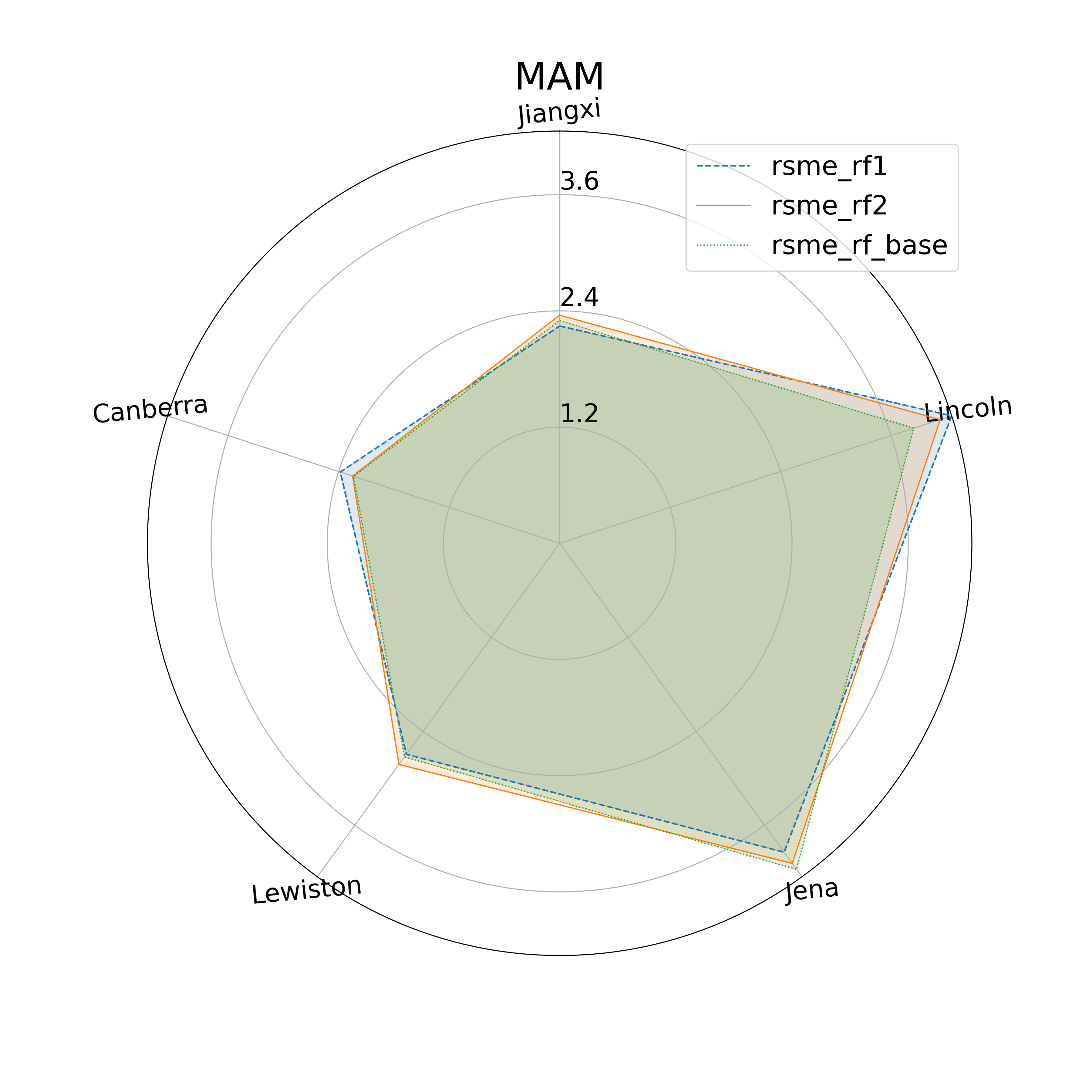}  \\
          \includegraphics[trim={0.6cm 0.6cm 0.6cm 0.6cm},clip,width=9cm]{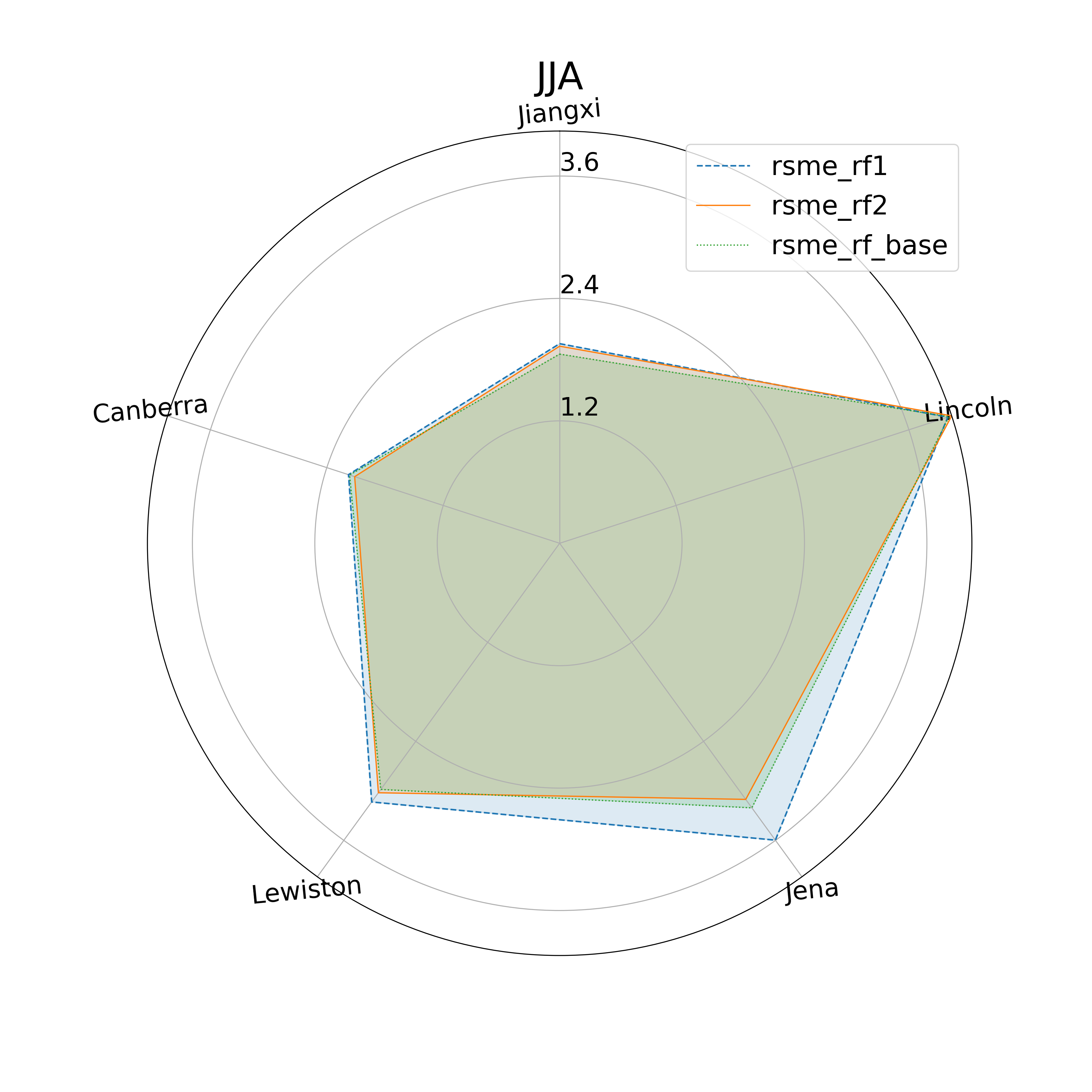} & \includegraphics[trim={0.6cm 0.6cm 0.6cm 0.6cm},clip,width=9cm]{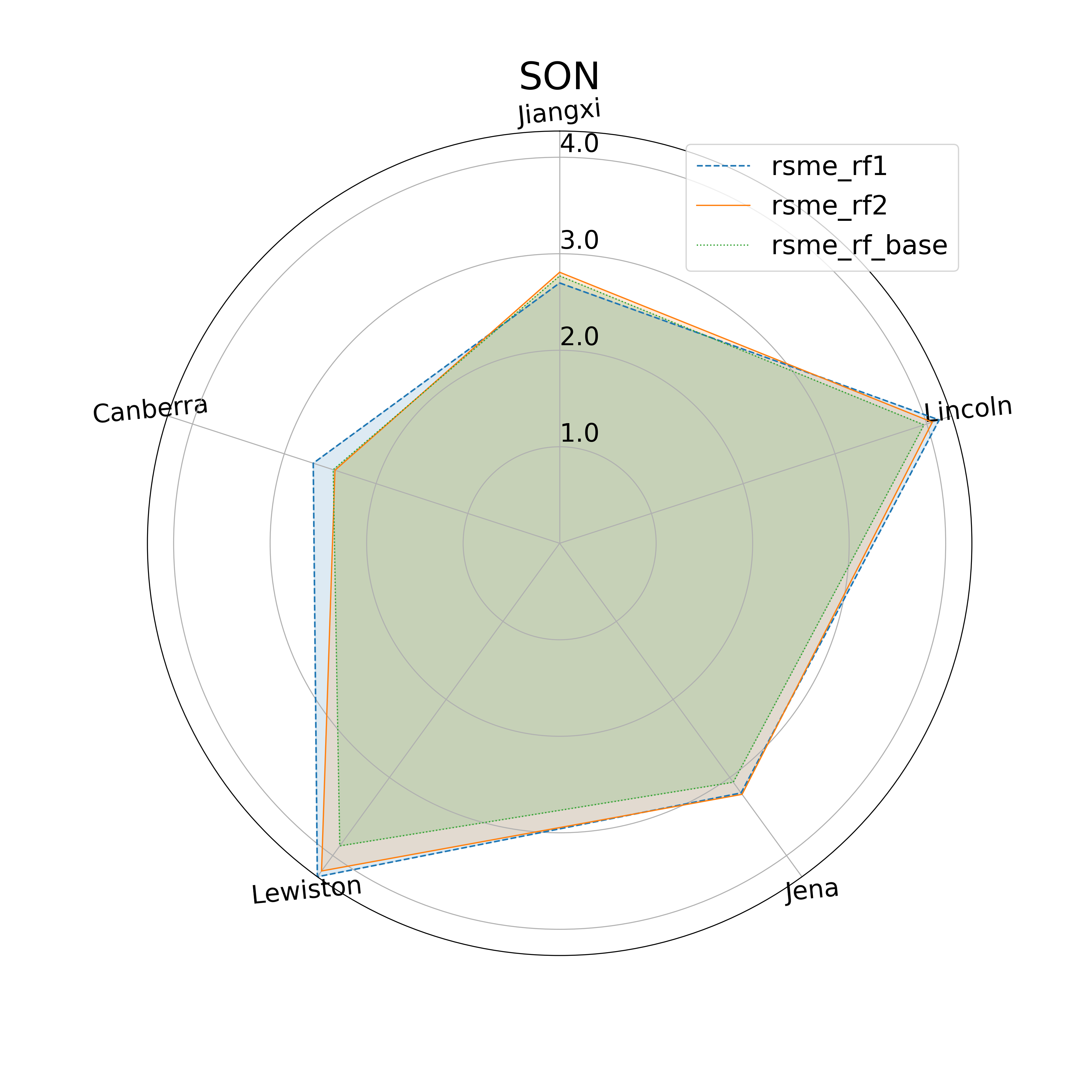}  \\
    \end{tabular}
 
   \caption{Summary radial graphs when $k$ = 10. In the radius we represent every location. Every season is one plot, as the plot title indicates: upper-left is DJF, upper-right is MAM, bottom-left is JJA and bottom-right is SON.}
    \label{fig:radials-k-10}
\end{figure*}

Figures  \ref{fig:radials-k-10} and  \ref{fig:radials-k-20} show the results with a smaller and higher number for parameter k. Looking at these plots, we can state that there is not a clear tendency, when we compare the three methods, it depends on the city and the season. However, as a general rule, the base case (learning with all the variables, in green) seems to provide smaller errors, while between the other two, in many cases, they perform quite similarly, and in others $rf_2$ wins (Fig. \ref{fig:radials-k-15} in season JJA for Lewiston, Lincoln and Jena) whereas sometimes $rf_1$ has lower error (same case for Jiangxi or Lewiston and Jena in MAM). Notice that the dotted line in blue corresponds to our ranking-based aggregation proposal, $rf_1$ (Alg. \ref{alg:exp}, line 14)  vs the solid line in orange, which corresponds to the top K selection by random forest importance, $rf_2$ (Alg. \ref{alg:exp}, line 15).

Observing the results when k=10 (Fig. \ref{fig:radials-k-10}) and k=15 (Fig. \ref{fig:radials-k-15}), the values differ, sometimes with larger differences, but the general analysis could be quite similar. These results show that Random Forest may be overfitting, which makes sense, as the number of instances it learns from is relatively small. In order to check the hypothesis of a possible overfitting, as Random Forest may need larger datasets to learn more robustly, we have included a simpler model into the experimentation. We chose Bayesian Ridge Regression \cite{box2011bayesian}, which is a variant of linear regression. It is a probabilistic model that introduces a Bayesian framework to estimate the parameters of the model. Even though it admits regularization parameters to prevent overfitting, we used the default case so that comparisons with Random Forest are fairer. Table \ref{tab:bay_base_rsme_all} shows the base results if we use this model, instead of Random Forest. In the experimental algorithm, it would just include the line \textsc{Learn\_Validate}(Bayesian\_Ridge,data) like in line 12 of Alg. \ref{alg:exp}.

 \begin{table}[]
    \centering

\begin{tabular}{lrrrr}
\toprule
location &     DJF &     MAM &     JJA &     SON \\
\midrule
 Jiangxi & 3.34568 & 2.34824 & 1.57824 & 2.71997 \\
 Lincoln & 5.71176 & 3.75950 & 3.70479 & 3.89288 \\
    Jena & 6.30099 & 4.07241 & 3.17720 & 3.17043 \\
Lewiston & 3.96669 & 2.78479 & 2.89577 & 3.98564 \\
Canberra & 1.91339 & 2.25791 & 2.02373 & 2.52020 \\
\bottomrule
\end{tabular}

\vspace*{0.25cm}
 \caption{Errors per location (rows) and season (columns) when using all the variables (Bayesian Ridge regressor).}
     \label{tab:bay_base_rsme_all}
 \end{table}

\begin{figure*}
    \centering
    \begin{tabular}{cc}
           \includegraphics[trim={0.25cm 0.25cm 0.25cm 0.25cm},clip,width=9cm]{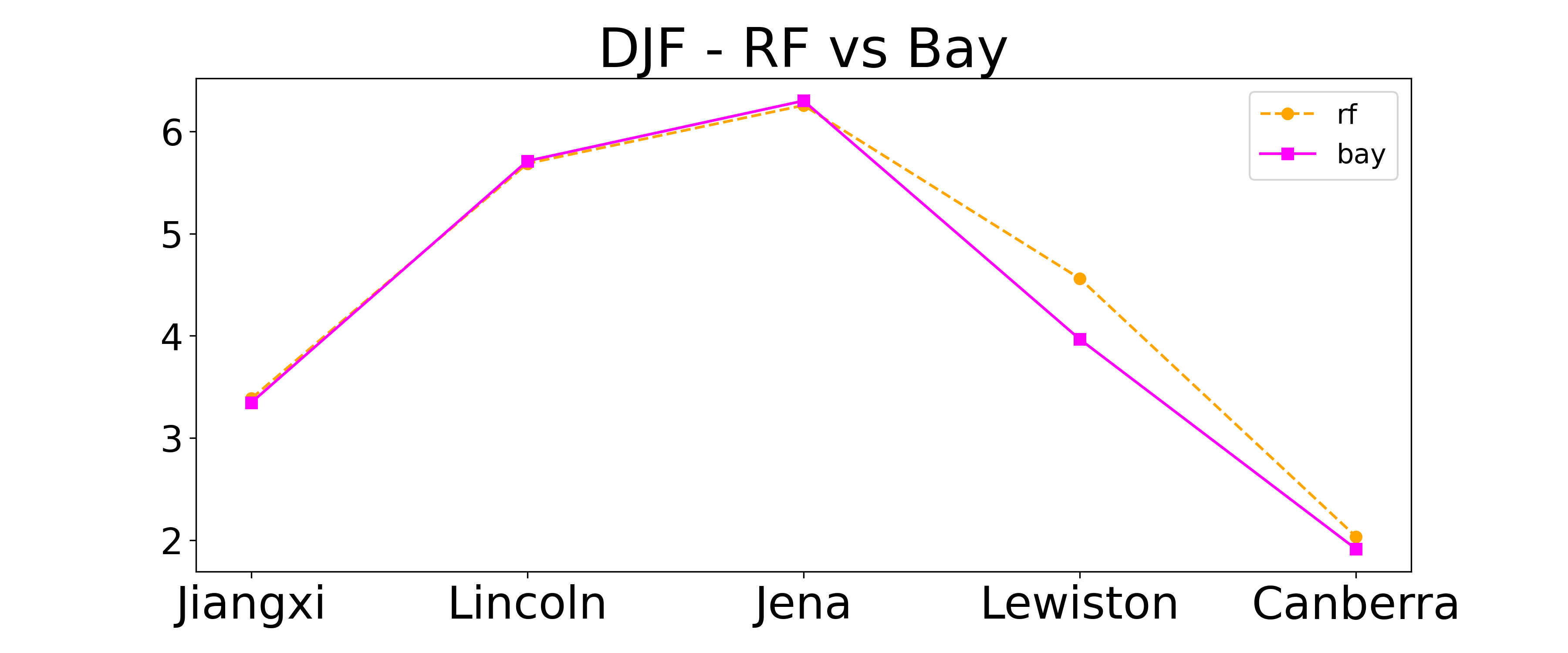} & \includegraphics[trim={0.25cm 0.25cm 0.25cm 0.25cm},clip,width=9cm]{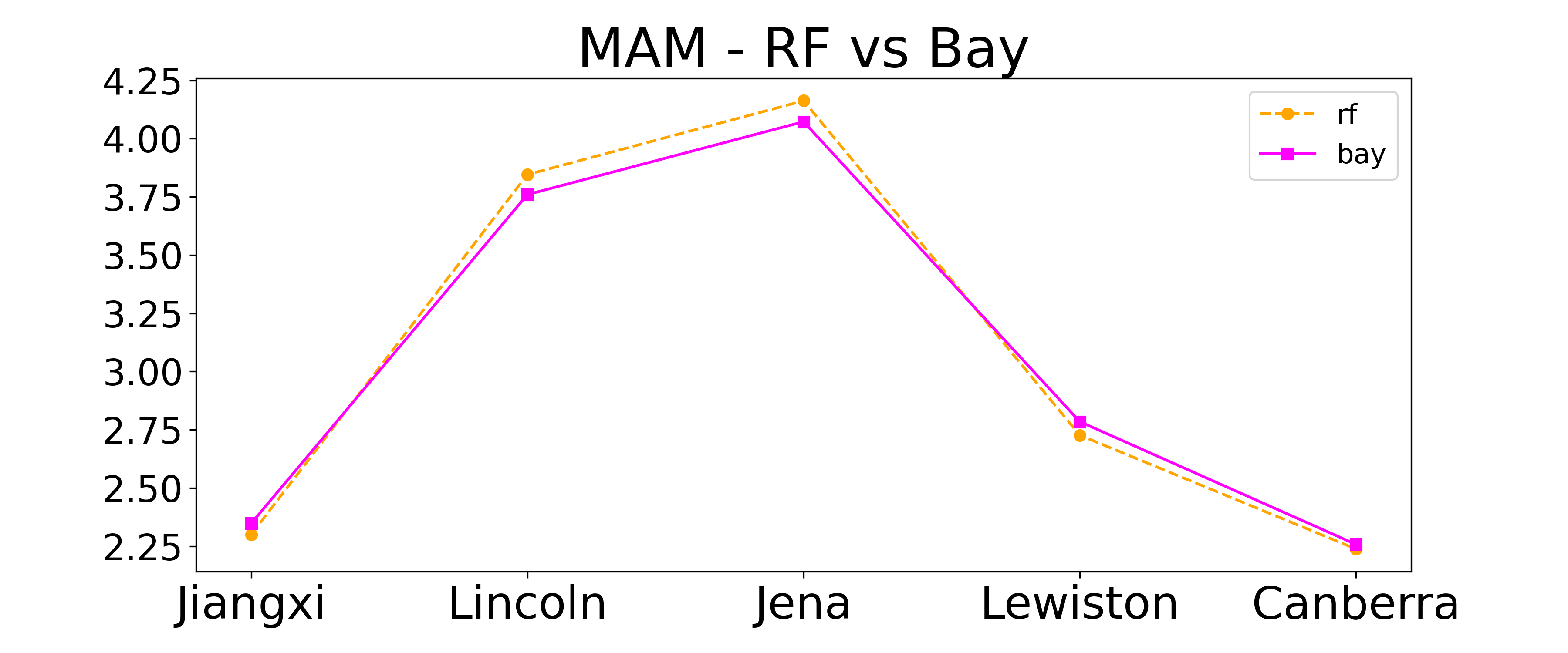}  \\
          \includegraphics[trim={0.25cm 0.25cm 0.25cm 0.25cm},clip,width=9cm]{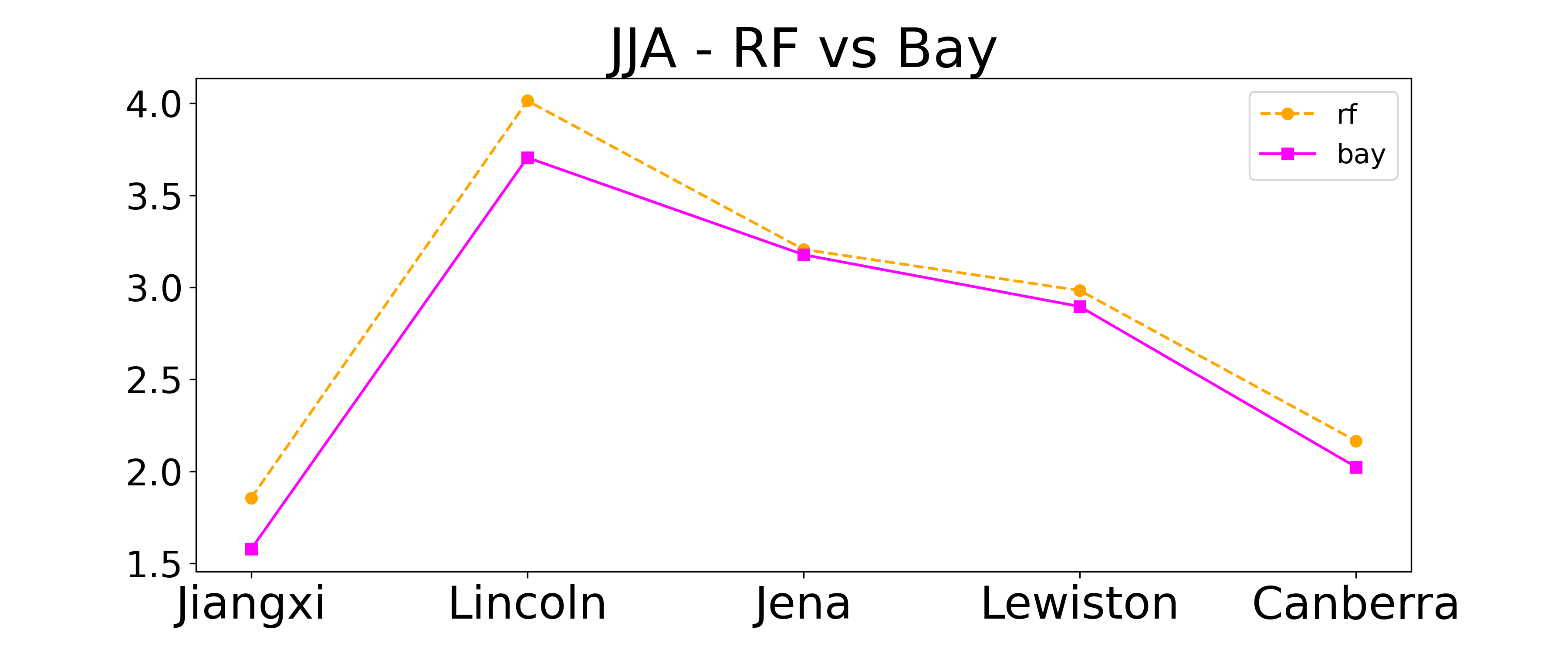} & \includegraphics[trim={0.25cm 0.25cm 0.25cm 0.25cm},clip,width=9cm]{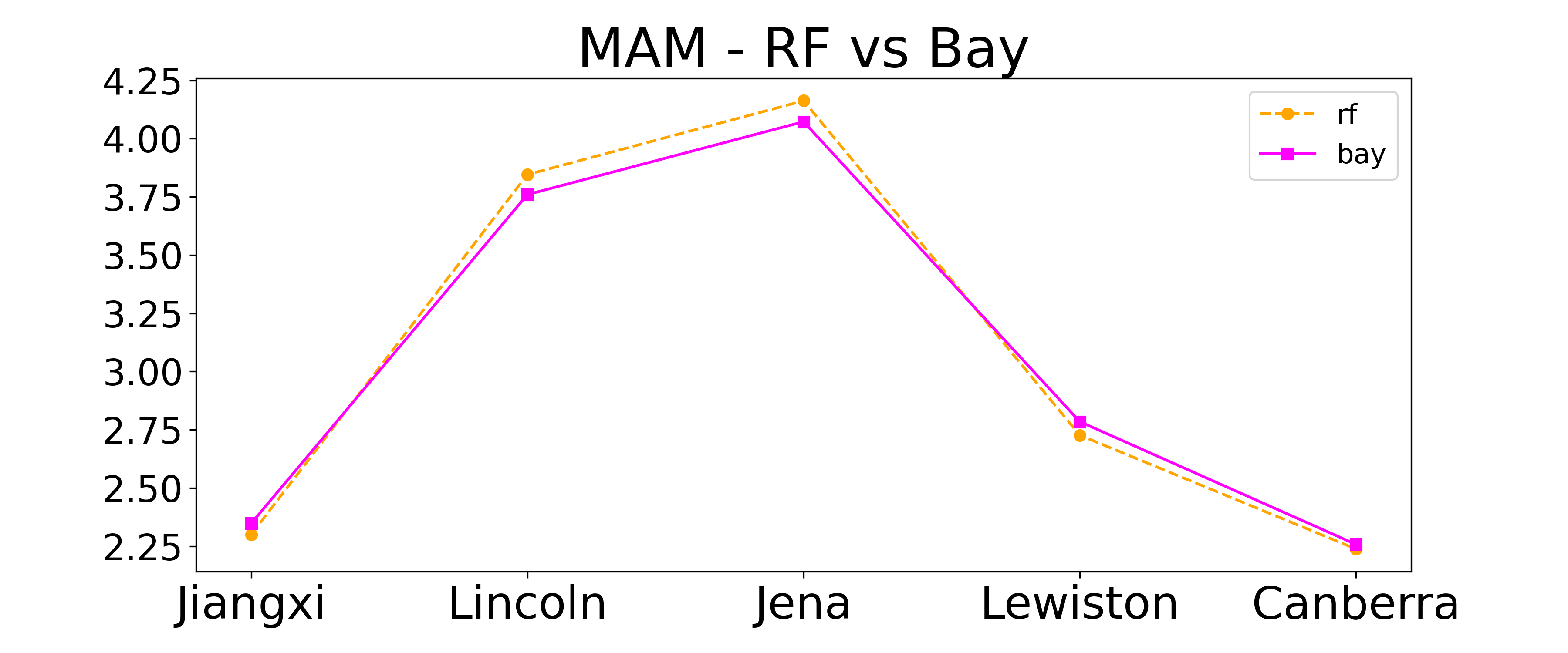}  \\
    \end{tabular}
 
    \caption{Comparing the rsme of the base models (all variables included) for models Random Forest (RF) and Bayesian Ridge (Bay).}
    \label{fig:rf-vs-bay}
\end{figure*}

To visually compare the reported values in Tables \ref{tab:rf_base_rsme_all} and \ref{tab:bay_base_rsme_all}, we have created the plots in Fig. \ref{fig:rf-vs-bay}. This comparison shows clearly that the performance of both methods is practically the same and, when there are difference, the linear regression provides lower error. That supports the hypothesis that, in the current framework, Random Forest may overfit.

\begin{figure*}
    \centering
    \begin{tabular}{cc}
           \includegraphics[trim={0.25cm 0.25cm 0.25cm 0.25cm},clip,width=9cm]{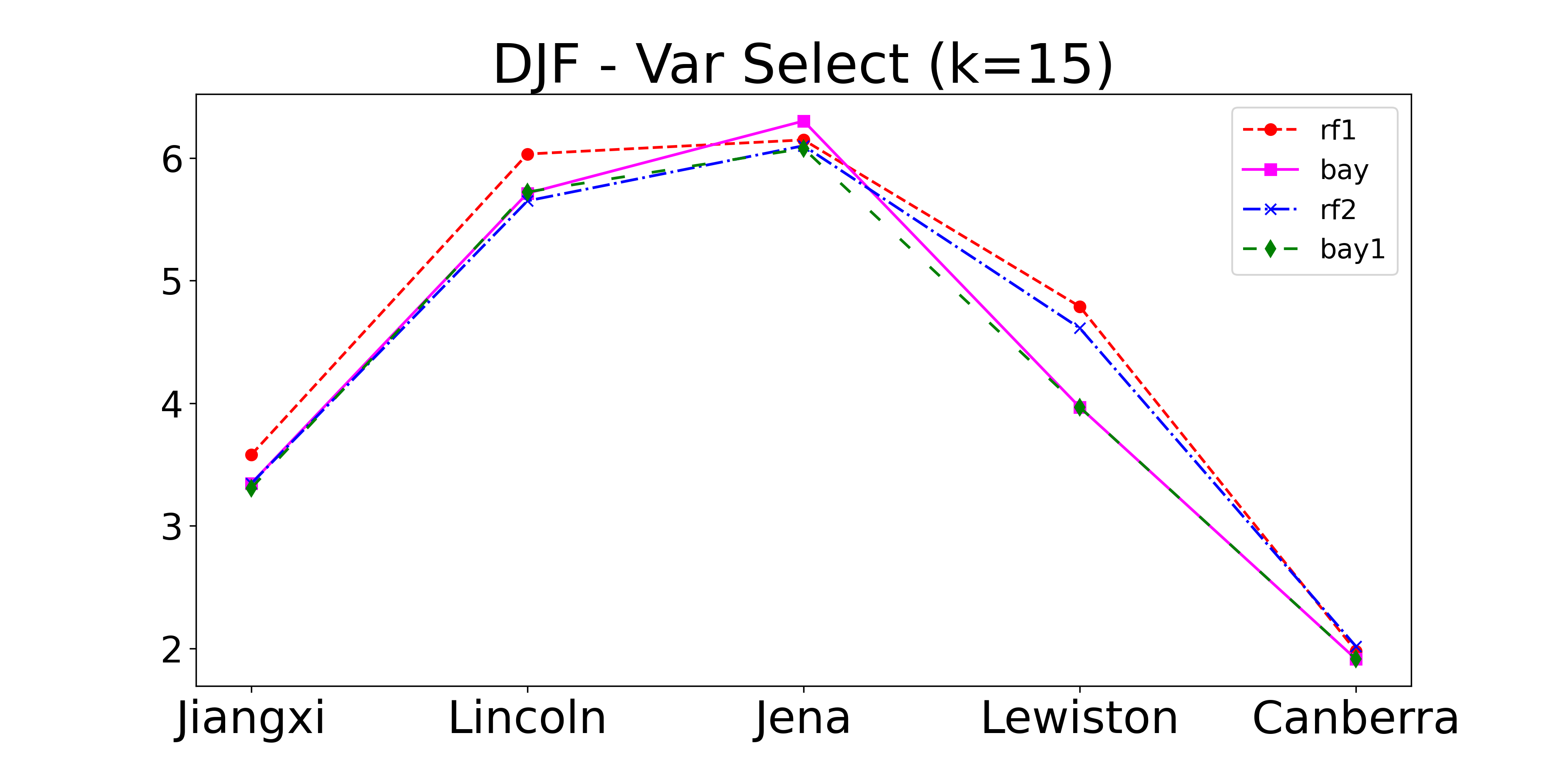} & \includegraphics[trim={0.25cm 0.25cm 0.25cm 0.25cm},clip,width=9cm]{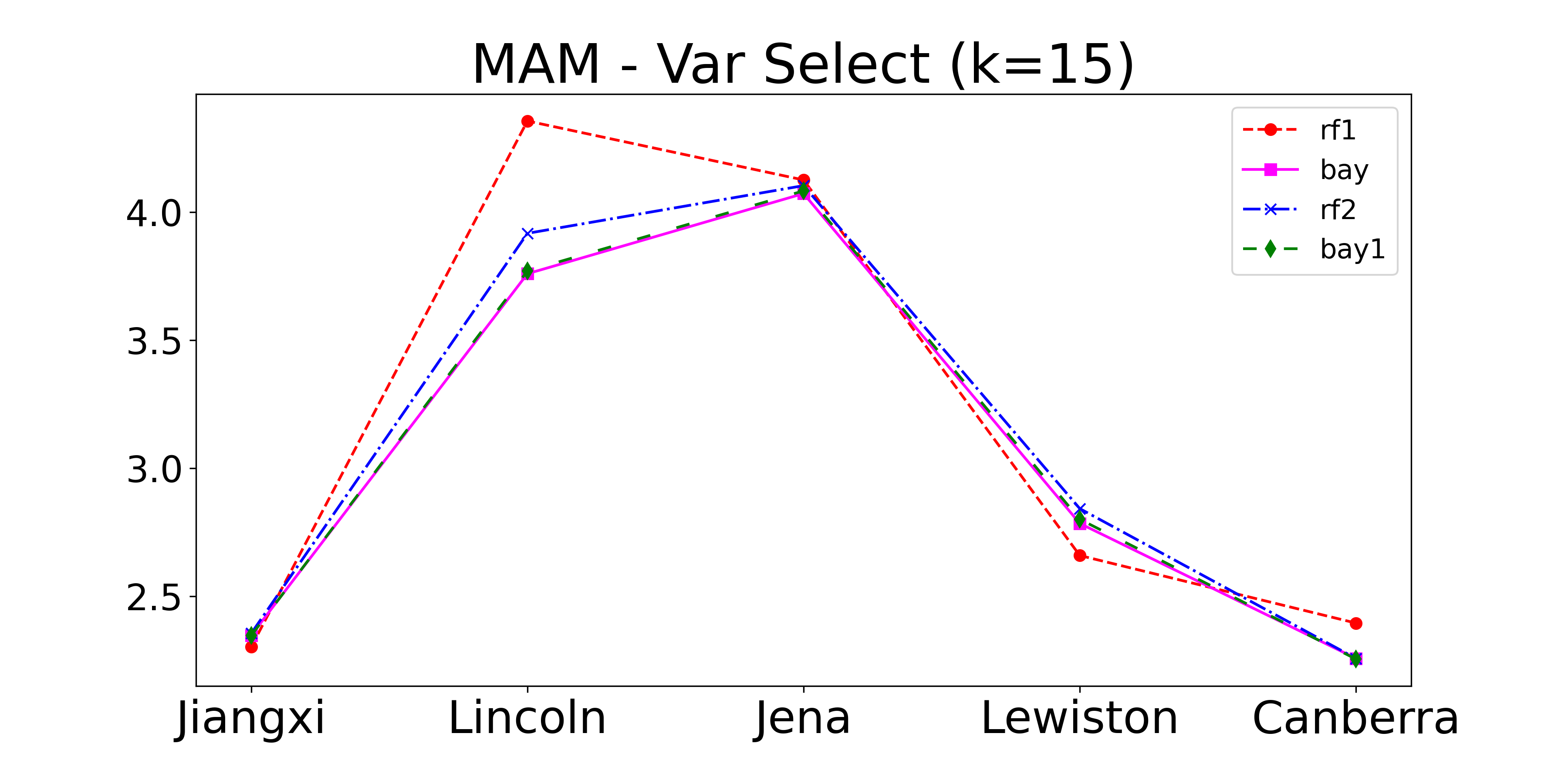}  \\
          \includegraphics[trim={0.25cm 0.25cm 0.25cm 0.25cm},clip,width=9cm]{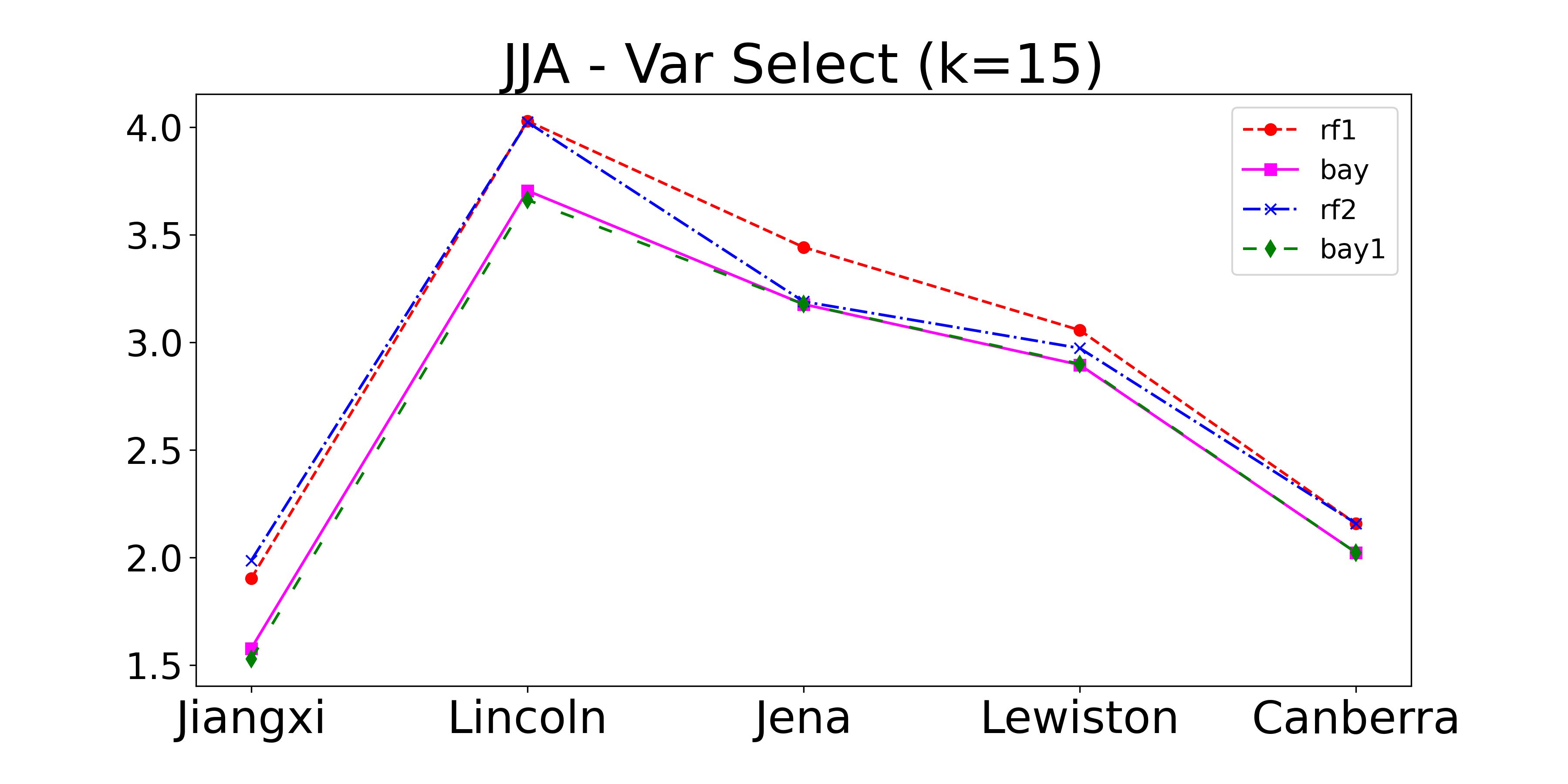} & \includegraphics[trim={0.25cm 0.25cm 0.25cm 0.25cm},clip,width=9cm]{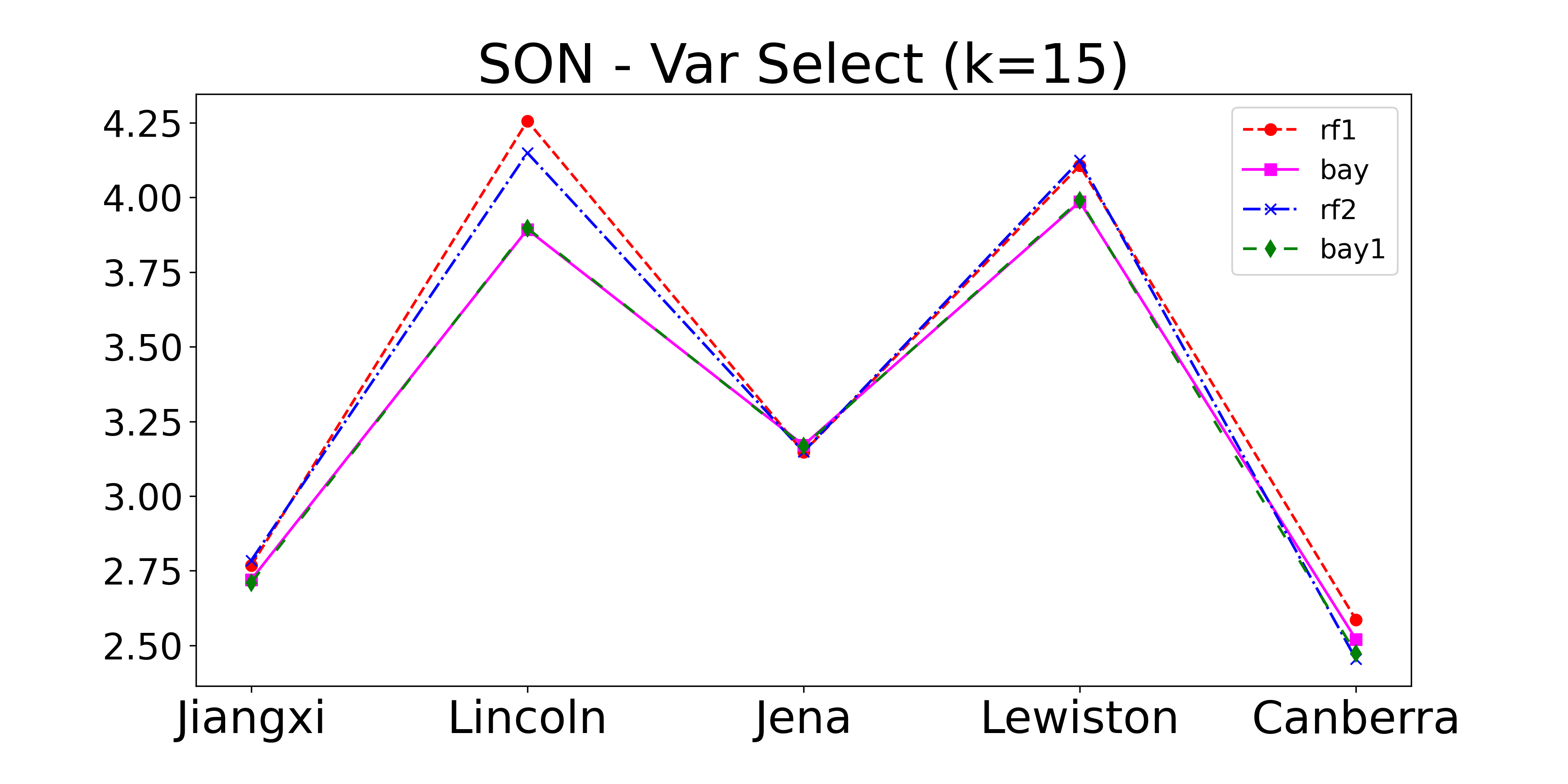}  \\
    \end{tabular}
 
    \caption{Comparing the rsme of the Bayesian Ridge model, when using all variables (bay), using the selection by aggregating rankings (bay1). As a reference, we also include the results of $rf_1$ and $rf_2$.}
    \label{fig:k-15-agg-bay}
\end{figure*}

As the primary purpose of this study is to evaluate the performance of the ranking aggregation to select the most significant variables, we will show the errors for the two cases where this ranking aggregation is used for selection, together with the base Bayesian model and the use of selecting the top k by the importance values of the RF ($rf_2$). We will use k=15, and the results are shown in Fig \ref{fig:k-15-agg-bay}.  From these plots, we can see how our aggregation method works much better using a simpler model as the Bayesian Ridge than when using a random Forest. We can also see that it doesn't always improve the performance of the base model, but it never affects its values strongly. 

\begin{figure*}
    \centering
    \begin{tabular}{cc}
           \includegraphics[trim={0.25cm 0cm 0.25cm 0.25cm},clip,width=9cm]{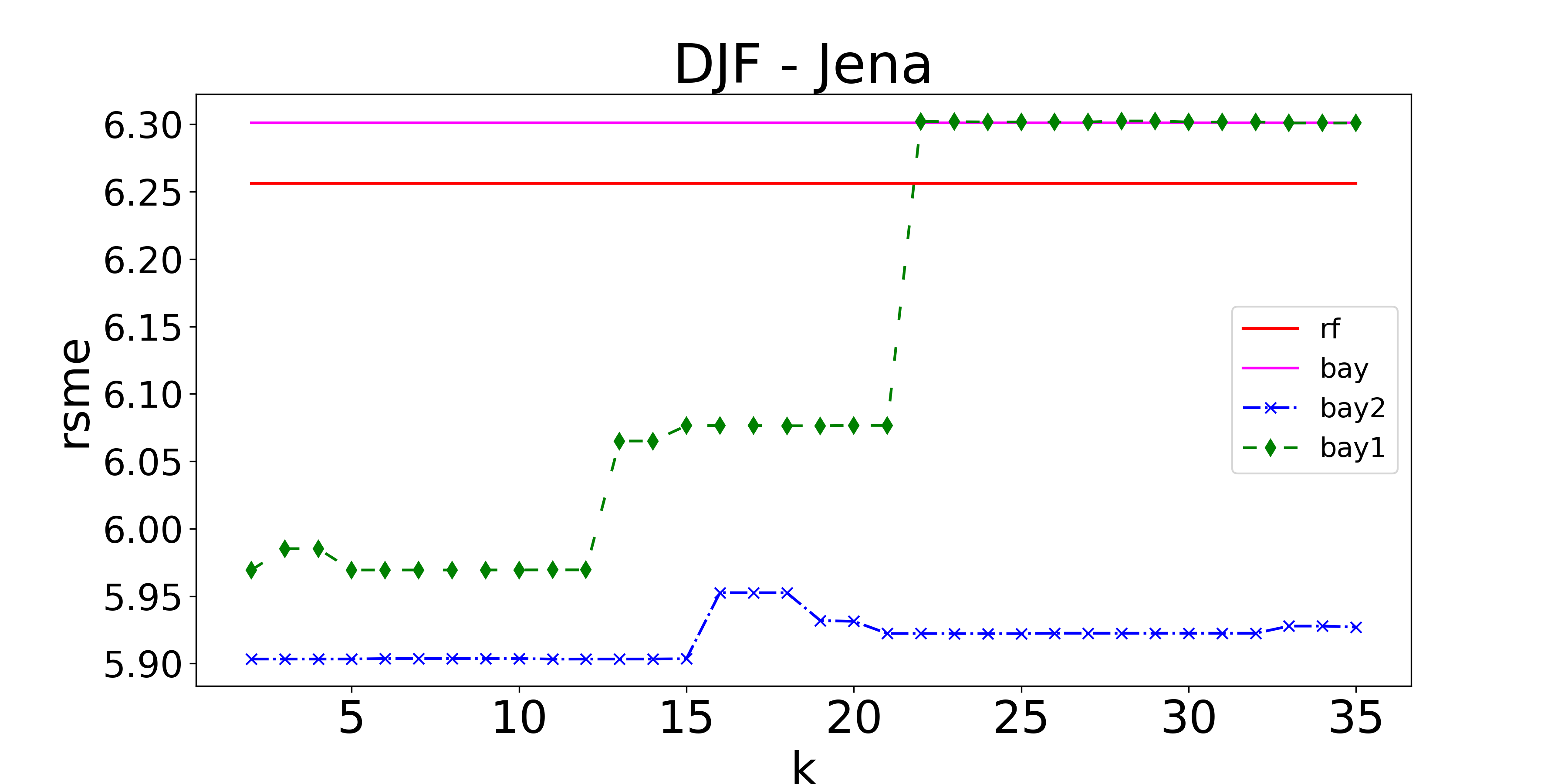} & \includegraphics[trim={0.25cm 0cm 0.25cm 0.25cm},clip,width=9cm]{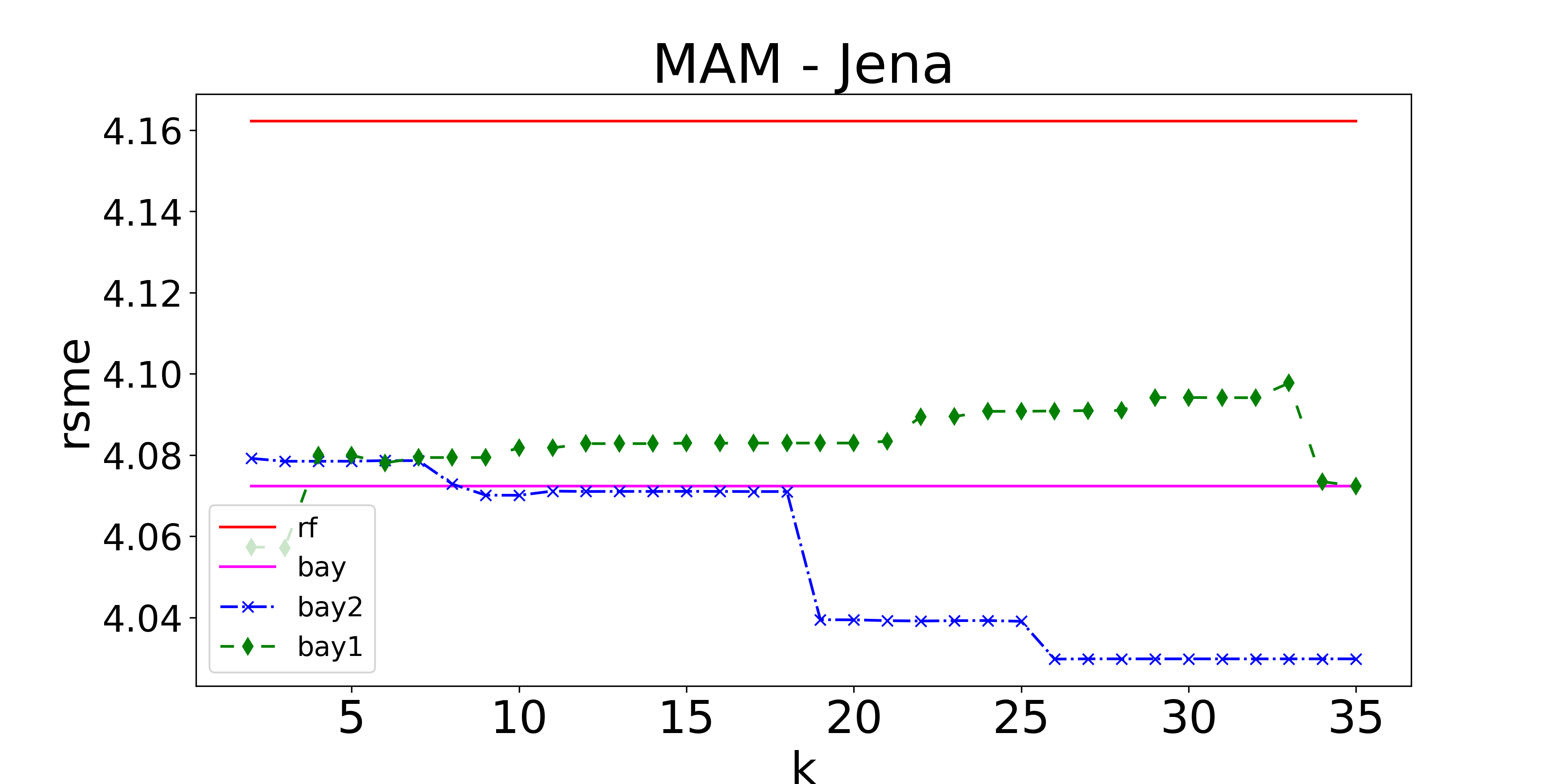}  \\
          \includegraphics[trim={0.25cm 0cm 0.25cm 0.25cm},clip,width=9cm]{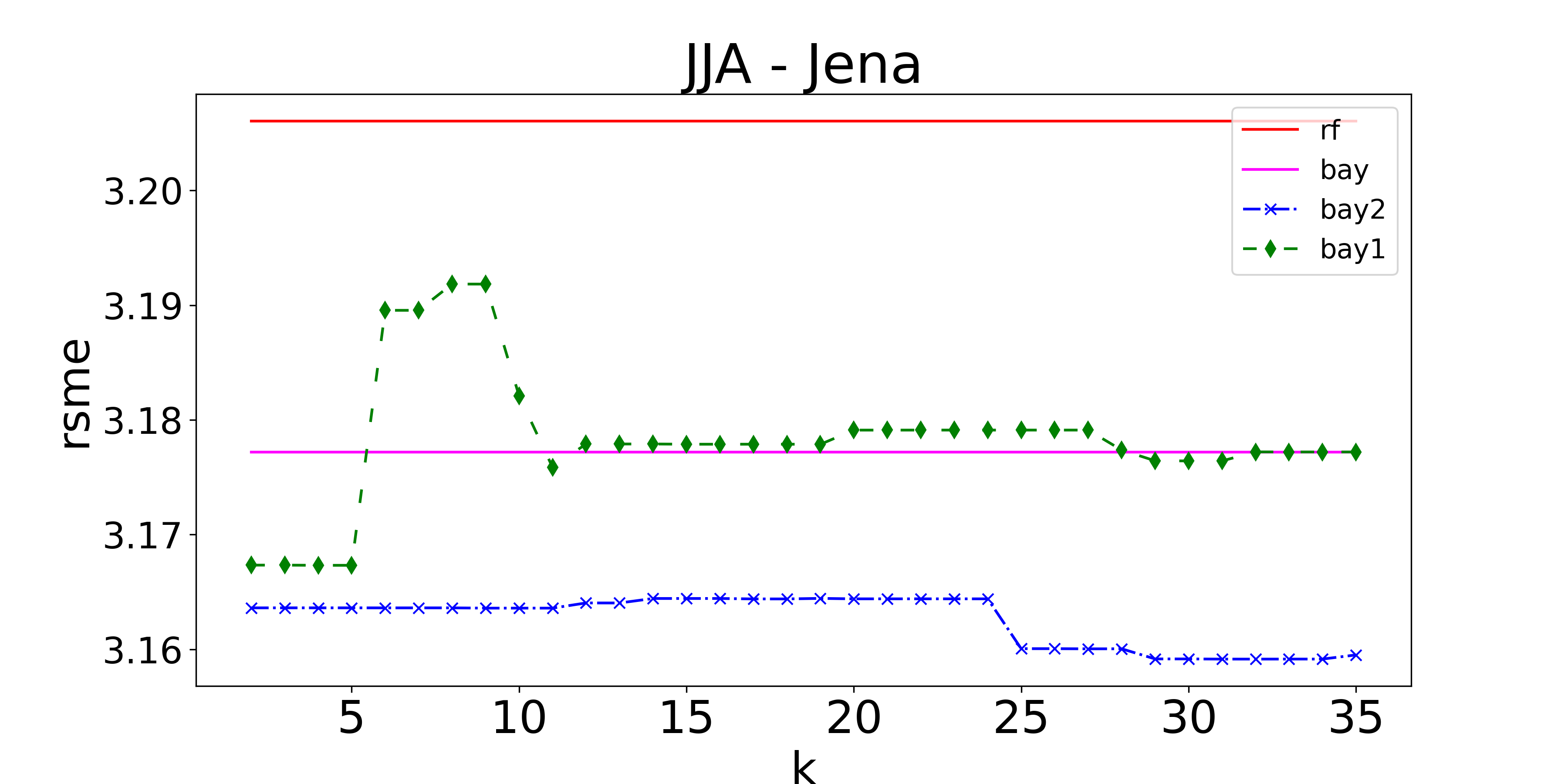} & \includegraphics[trim={0.25cm 0cm 0.25cm 0.25cm},clip,width=9cm]{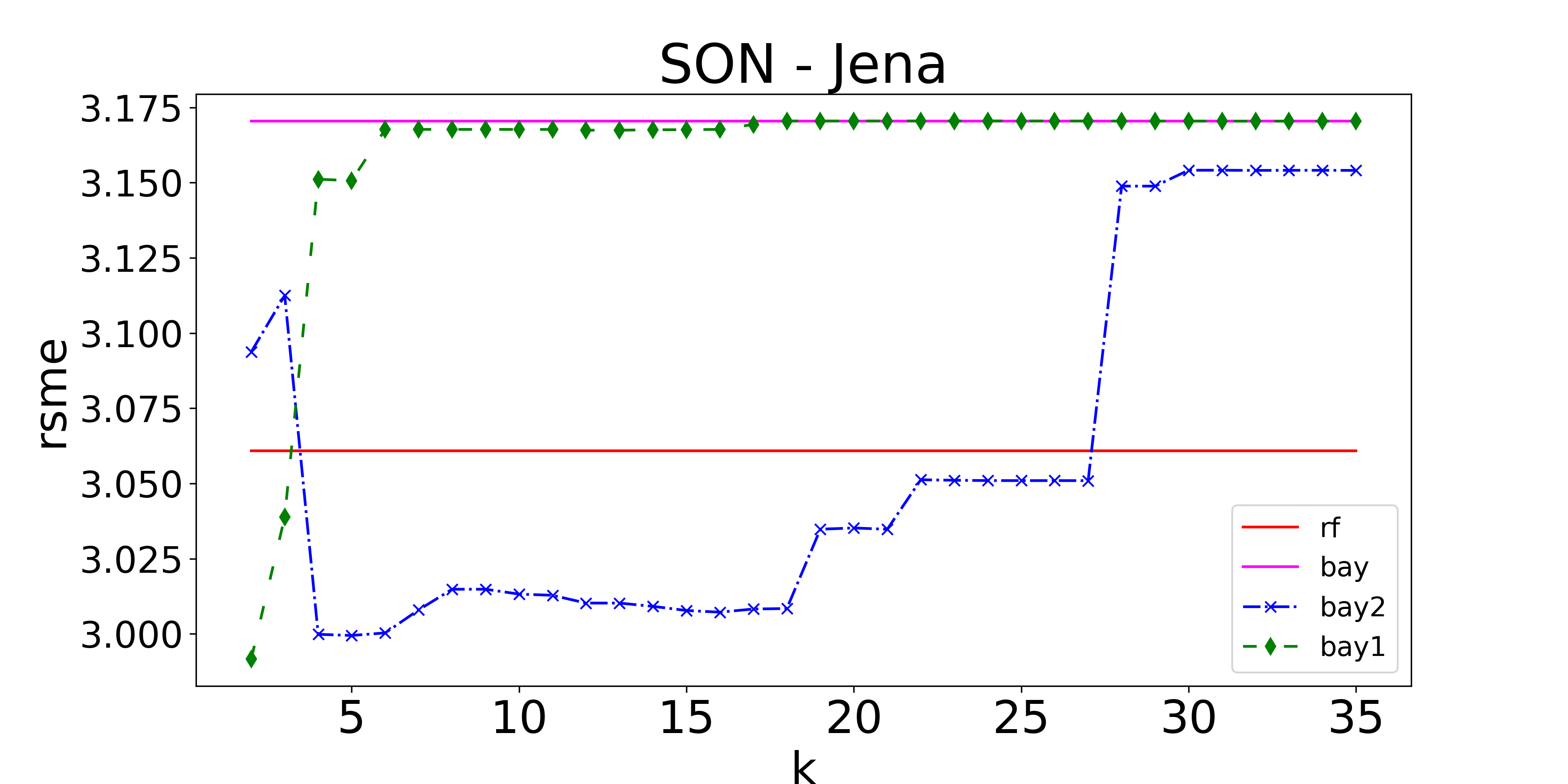}  \\
    \end{tabular}
 
    \caption{Comparing the aggregation selection and the rf selector for Bayesian Ridge and city Jena. $rf$ shows the error for the base RF as a reference. In the horizontal axis: the value K. }
    \label{fig:all-ks-Jena}
\end{figure*}

\begin{figure*}
    \centering
    \begin{tabular}{cc}
           \includegraphics[trim={0.25cm 0cm 0.25cm 0.25cm},clip,width=9cm]{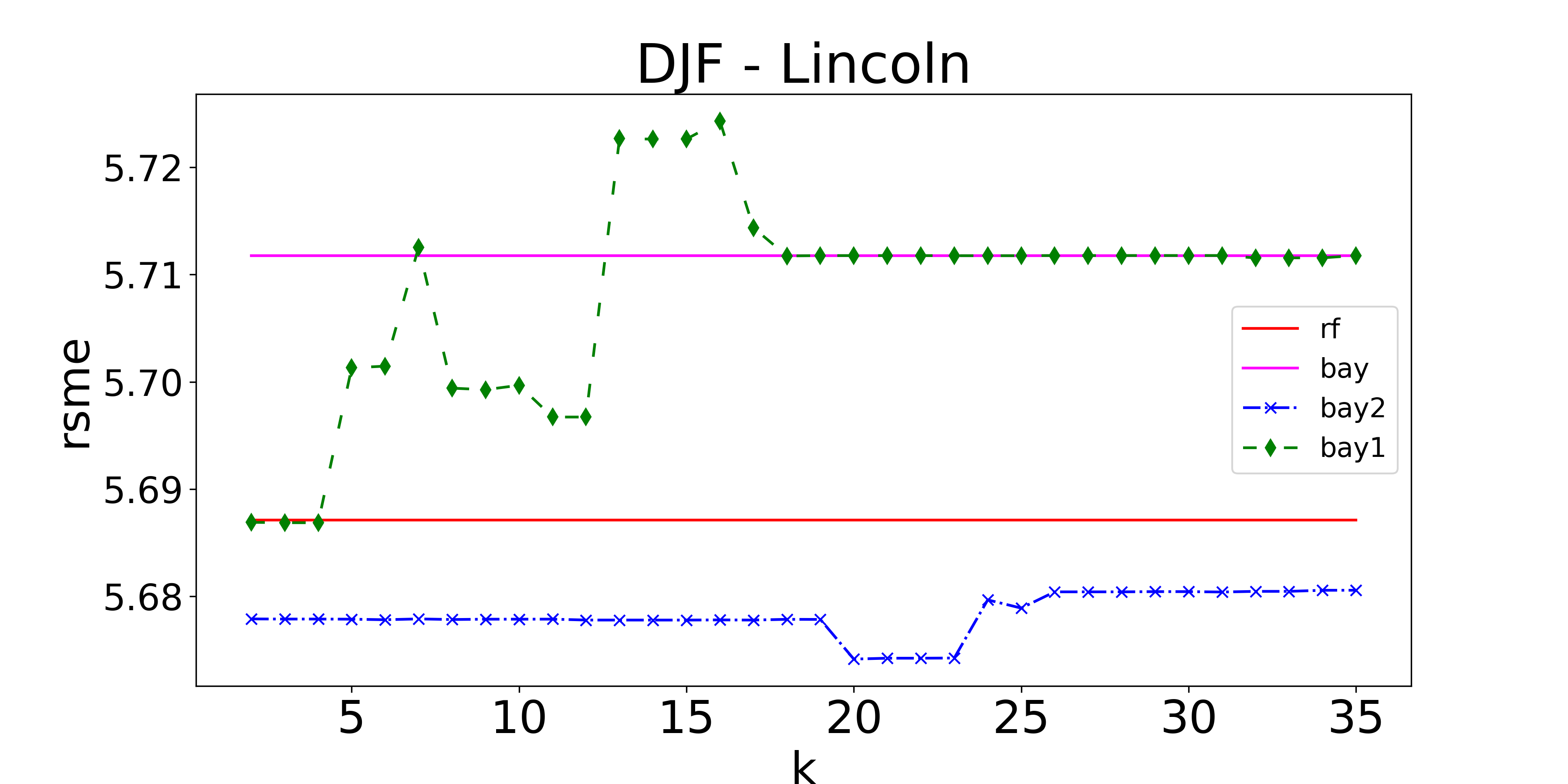} & \includegraphics[trim={0.25cm 0cm 0.25cm 0.25cm},clip,width=9cm]{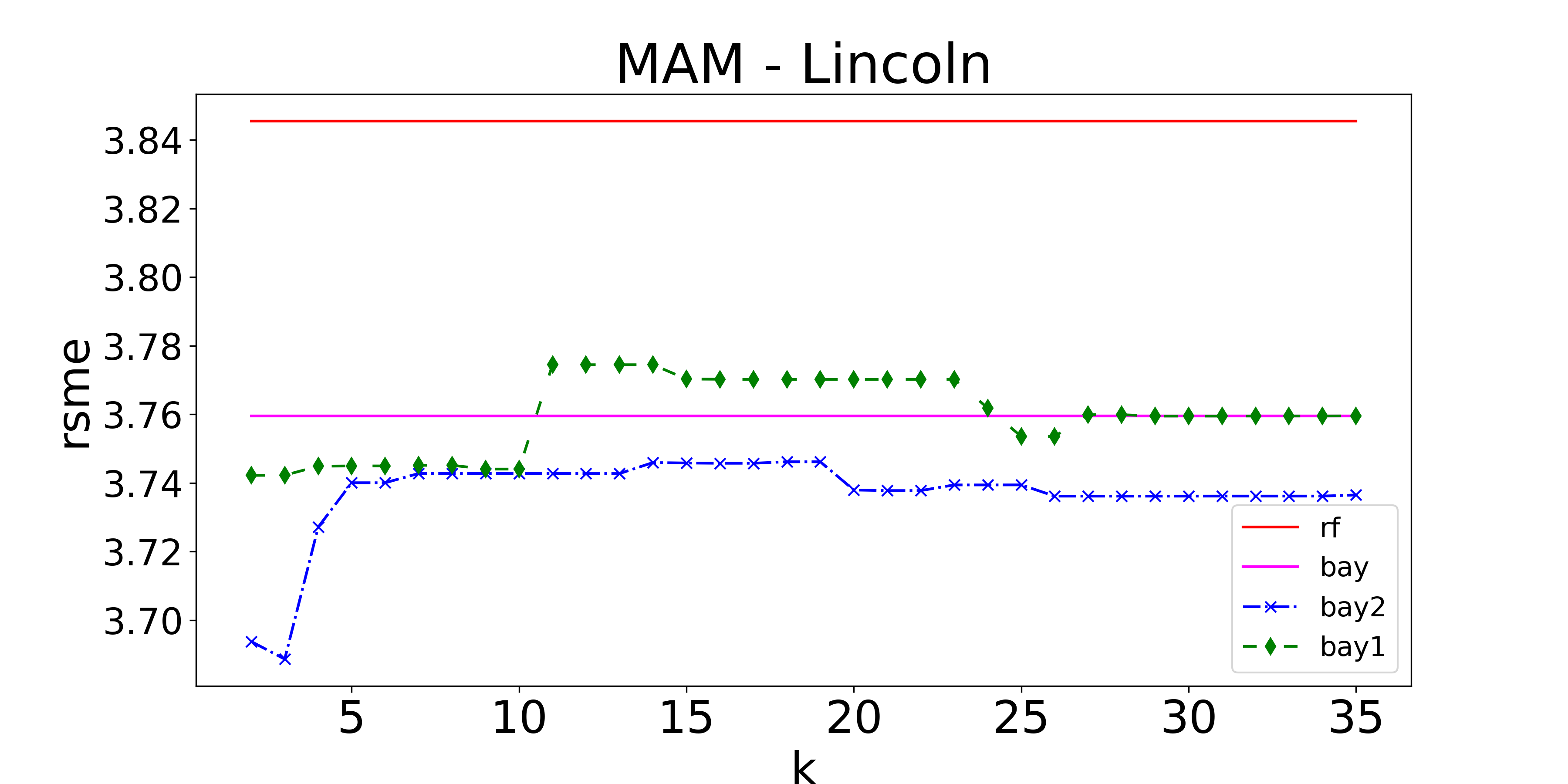}  \\
          \includegraphics[trim={0.25cm 0cm 0.25cm 0.25cm},clip,width=9cm]{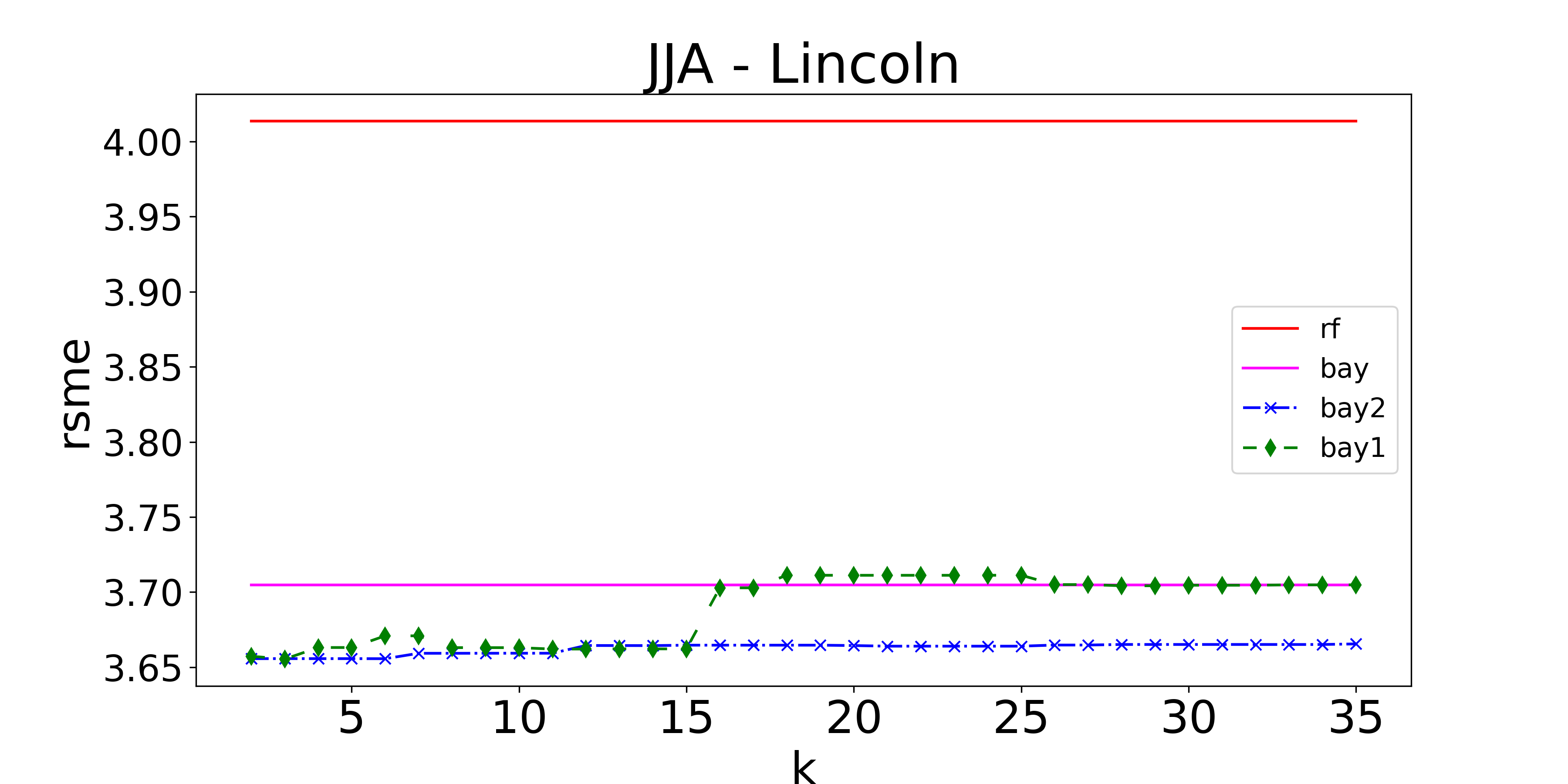} & \includegraphics[trim={0.25cm 0cm 0.25cm 0.25cm},clip,width=9cm]{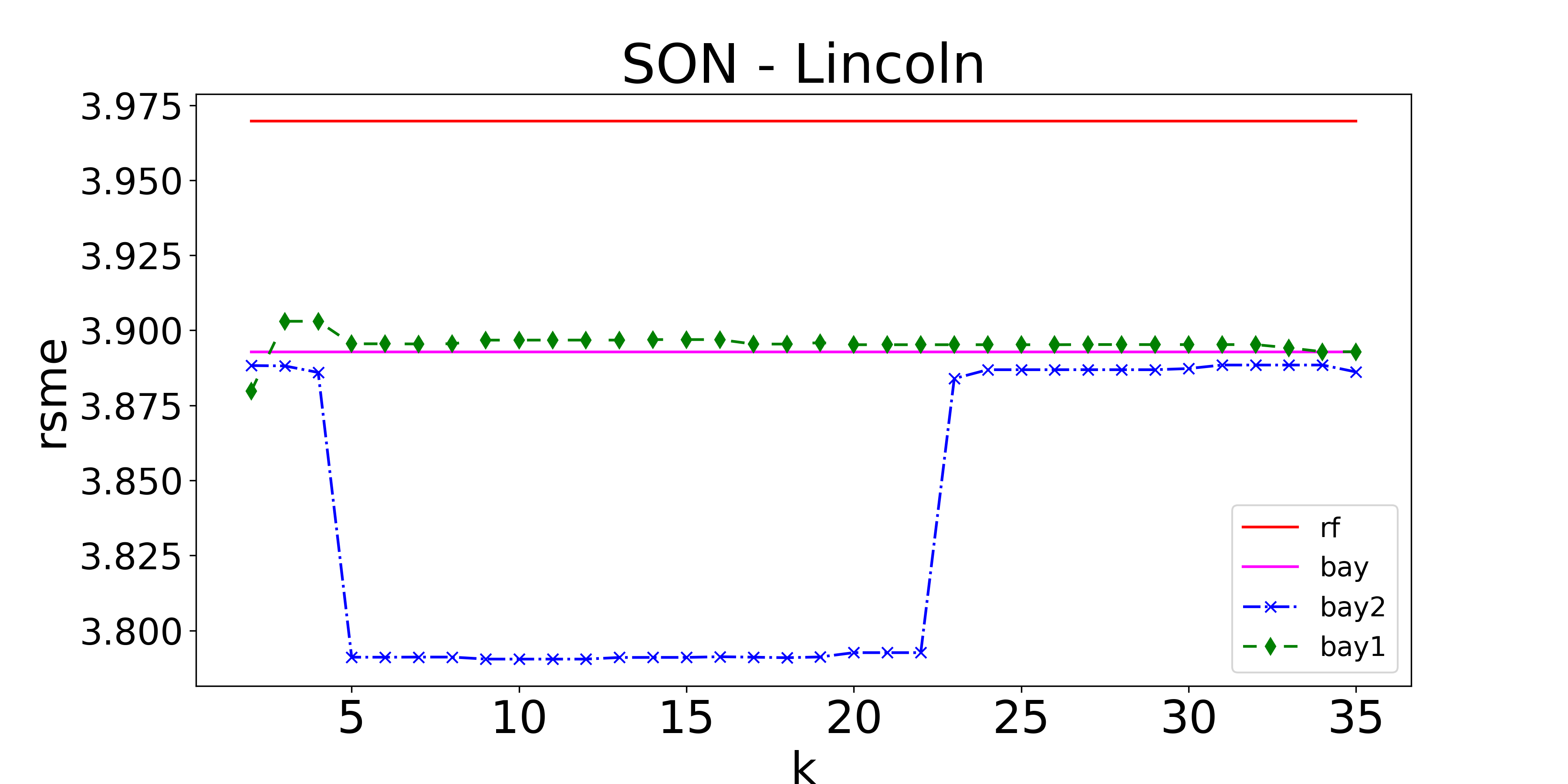}  \\
    \end{tabular}
 
    \caption{Comparing the aggregation selection and the rf selector for Bayesian Ridge and city Lincoln. $rf$ shows the error for the base RF as a reference.  In the horizontal axis: the value K.}
    \label{fig:all-ks-Lincoln}
\end{figure*}

As Random Forest also provides a selection method, one fair comparison would be applying the selection of the top k features discovered by the importance feature technique, and learn a Bayesian Ridge model. Following the previous notation, it would be: $bay_2$ $\leftarrow$ \textsc{Learn\_Validate}(Bayesian\_Ridge,data[$sel\_rf$]). In this case, and in order to also explore the distinct values for k, we will plot in the x-axis this value for k, and in the y-axis the rsme values for applying our aggregation technique versus RF selection. In order to keep under control the number of figures, only two locations whose values resulted more variable in the previous exploration have been chosen: Jena and Lincoln. Figures \ref{fig:all-ks-Jena} and \ref{fig:all-ks-Lincoln} shows, per every season, how these two selection techniques behave with respect to the rsme in the Bayesian Regresson model. When looking at the graphs the selection based on RF importance features provides generally lower rsme error. However, when looking at the absolute difference, it goes in an interval of [0,0.1] degrees. With respect to the ideal value for $k$, we see that there could be distinct behaviors depending on the location and the season. In these cases a value between 10 and 20 could be reasonable, as the performance is good enough, almost as good as using all the variables, and the models will generalize better.  

It could be relevant how much the three correlation rankings correlate in every case and their correlation with respect to the borda aggregation. We show Table \ref{fig:rankings-corr-per-location-and-season} for that analysis. In it, the column rankings shows the correlation of the ranks from Pearson, Spearman and Kendall correlation. This value is the average of the three pairwise correlations. The column borda reports the average of computing the correlation of the aggregated ranking with respect to the original three. The data are shown per location and season. Considering these two dimensions, the global mean values are reported on the right column and the two last rows. This information is really quite valuable to see the concordance of the three metrics, and we can indeed see why Lewiston is sensitive to the distinct regressor models (see Figures \ref{fig:rf-vs-bay} and \ref{fig:k-15-agg-bay}; in this case the distribution at Fig. \ref{fig:lw_winter_pairplot} can also be related). We can see that it is the one whose rankings have a lower correlation in DJF and JJA, but also on average. When the correlation is lower than 0.75, it is clear that the errors can oscillate more. So, this is one factor to be studied if a global scheme is included when studying the importance of the variables. Finally, as it is clear and expected, the aggregate ranking (borda) increases the correlation with respect to the one computed fir initial three input rankings. The best two values correspond to Lincoln and Jena, and those are the cases we plotted by K values (Figures \ref{fig:all-ks-Jena} and \ref{fig:all-ks-Lincoln}). It is remarkable how in most seasons a value of K=20 (or smaller) get the same performance as using the whole set of variables. 

\begin{table}[]
\begin{tabular}{ll|rrrr||c}
\toprule
{} & corr &      DJF &      MAM &      JJA &      SON &   Avg \\
Location &             &          &          &          &          &           \\
\midrule
Jiangxi  &     ranking &  0.73860 &  0.79020 &  0.79800 &  0.76230 &  0.772275 \\
  &       borda &  0.84040 &  0.87540 &  0.88290 &  0.84840 &  0.861775 \\
Jena     &     ranking &  0.80040 &  0.83110 &  0.83920 &  0.88150 &  0.838050 \\
    &       borda &  0.87920 &  0.89830 &  0.90410 &  0.93290 &  0.903625 \\
Lincoln  &     ranking &  0.87030 &  0.74810 &  0.84340 &  0.86080 &  0.830650 \\
  &       borda &  0.92230 &  0.84740 &  0.90660 &  0.91490 &  0.897800 \\
Lewiston &     ranking &  0.65500 &  0.84760 &  0.73970 &  0.81590 &  0.764550 \\
 &       borda &  0.79340 &  0.90450 &  0.84050 &  0.89210 &  0.857625 \\
Canberra &     ranking &  0.81510 &  0.82890 &  0.70770 &  0.82650 &  0.794550 \\
 &       borda &  0.89390 &  0.89480 &  0.81620 &  0.89700 &  0.875475 \\ \\ \hline
Avg  &     ranking &  0.77588 &  0.80918 &  0.78560 &  0.82940 &  0.800015 \\
  &       borda &  0.86584 &  0.88408 &  0.87006 &  0.89706 &  0.879260 \\
\bottomrule
\end{tabular}\vspace*{0.25cm}
  \caption{Concordance/correlation of the rankings.}
    \label{fig:rankings-corr-per-location-and-season}
\end{table}


\subsection{Evaluation of the results}

After the provided experimentation and all the results we have generated, there are some points to be highlighted:

\begin{itemize}
    \item Our ranking-based technique has proven to be helpful when applying it as a feature subset selection technique. It works quite similarly as rf feature importance, but it presents some advantages: (1) it is not model dependant, (2) this is a basic scheme that can be improved by integrating other metrics or correlations and/or by using other aggregation methods (distinct from borda could be studied)
    \item Every location and season presents its particularities, which this framework could extract by studying the distinct metrics.
    \item For a single cell a simple model as a linear regression provides better errors than Random Forest. 
\end{itemize}

The exploration, experimentation and the plotted results have contributed to the main purpose look for a Machine Learning technique accounting for rankings to improve the forecasting process.

\section{Conclusions and further work}
\label{sec:conclusions-further-work}

The work presented here is an initial study whose final aim is to provide insight into which variables affect the most error when performing temperature forecasts. In this paper, we have shown that the ranking-based methodology works for this purpose. The five cases of use have shown that our methodology successfully reduces the number of variables, which seems to be meaningful enough because the errors are close to the base model (all variables). It would be a key point to determine some parameters, such as the best k value or which aggregation to use. This framework has been shown with two models, but it can be extended to any regressor that is available. 

Random Forest has been proven to be a very powerful model, and it also allows explainability methods, as explained in the paper. However, when performing this study at a single-cell level, the results are not highly reliable due to overfitting. We could try to alleviate that problem in future work by using bigger areas of N x N instead of a single cell (900 weeks or data points, that makes around 225 weeks per season). For example, to analyze the area of Jena, if N=3, we would use as input the data the corresponding adjacent cells, and we will multiply by 9 the number of instances to learn. 

As additional future work, we should also investigate the pairwise relationships between features, as the pairplots showed some of them are strongly related. If we can establish redundant or dependant variables, or even groups of variables, by clustering, we could avoid the system detecting the same information repeated by selecting variables expressing the same information or the existence of confounding factors. This is a very promising line when analyzing the impact of Earth variables in the forecast errors. Finally, we could also focus on special events, such as extreme temperature values, which are the most important errors to avoid. 


\bibliographystyle{IEEEtran}
\bibliography{ml-fore}

\end{document}